\documentclass[ijoc,sglanonrev]{informs4}
\usepackage{eqndefns-left} 
\RequirePackage{tgtermes}
\RequirePackage{newtxtext}
\RequirePackage{newtxmath}
\RequirePackage{bm}
\RequirePackage{endnotes}
\usepackage{bbm}
\usepackage{graphicx}
\usepackage[colorlinks=true, allcolors=blue]{hyperref}
\usepackage{multirow}
\usepackage{array}
\usepackage{amsfonts}
\usepackage{booktabs}
\usepackage{tabularx} 
\usepackage{tikz}
\usepackage{subfigure}
\usepackage[flushleft]{threeparttable}
\usepackage{tabularray}

\newcommand*\circled[1]{\tikz[baseline=(char.base)]{
		\node[shape=circle,draw,inner sep=2pt] (char) {#1};}}

\newsavebox\CBox

\newcolumntype{L}[1]{>{\raggedright\let\newline\\\arraybackslash\hspace{0pt}}m{#1}}
\newcolumntype{C}[1]{>{\centering\let\newline\\\arraybackslash\hspace{0pt}}m{#1}}
\newcolumntype{R}[1]{>{\raggedleft\let\newline\\\arraybackslash\hspace{0pt}}m{#1}}

\appto\TPTdoTablenotes{\footnotesize}

\OneAndAHalfSpacedXII 

\usepackage{algorithm}
\usepackage{algpseudocode}
\usepackage{tikz}

\usepackage{natbib}
 \bibpunct[, ]{(}{)}{,}{a}{}{,}%
 \def\bibfont{\small}%
\EquationsNumberedThrough    

\TheoremsNumberedThrough     
\ECRepeatTheorems  %

\MANUSCRIPTNO{IJOC-0001-2024.00}

\begin{document}


\RUNAUTHOR{Dong et. al.}

\RUNTITLE{Causally-Aware Spatio-Temporal Multi-Graph Convolution Network for Accurate and Reliable Traffic Prediction}

\TITLE{\Large Causally-Aware Spatio-Temporal Multi-Graph Convolution Network for Accurate and Reliable Traffic Prediction}

\ARTICLEAUTHORS{%


\AUTHOR{Pingping Dong}
\AFF{Department of Industrial and Systems Engineering, The Hong Kong Polytechnic University, Kowloon, Hong Kong; School of Economics and Management, Tongji University, Siping Road 1239, Shanghai 200092, China.}

\AUTHOR{Xiao-Lin Wang}
\AFF{Business School, Sichuan University, Chengdu 610065, China.}

\AUTHOR{Indranil Bose}
\AFF{Department of Information Systems, Supply Chain Management \& Decision Support, NEOMA Business School, Reims, 51100, France.}

\AUTHOR{Kam K.H. Ng}
\AFF{Department of Aeronautical and Aviation Engineering, The Hong Kong Polytechnic University, Kowloon, Hong Kong.}

\AUTHOR{Xiaoning Zhang}
\AFF{School of Economics and Management, Tongji University, Siping Road 1239, Shanghai 200092, China.}

\AUTHOR{Xiaoge Zhang$^*$
}
\AFF{Department of Industrial and Systems Engineering, The Hong Kong Polytechnic University, Kowloon, Hong Kong.}
} 

\ABSTRACT{%
Accurate and reliable prediction has profound implications to a wide range of applications, such as hospital admissions, inventory control, route planning. In this study, we focus on an instance of spatio-temporal learning problem\textemdash traffic prediction\textemdash to demonstrate an advanced deep learning model developed for making accurate and reliable forecast. Despite the significant progress in traffic prediction, limited studies have incorporated both explicit (e.g., road network topology) and implicit (e.g., causality-related traffic phenomena and impact of exogenous factors) traffic patterns simultaneously to improve prediction performance. Meanwhile, the variability nature of traffic states necessitates quantifying the uncertainty of model predictions in a statistically principled way; however, extant studies offer no provable guarantee on the statistical validity of confidence intervals in reflecting its actual likelihood of containing the ground truth. In this paper, we propose an end-to-end traffic prediction framework that leverages three primary components to generate accurate and reliable traffic predictions: dynamic causal structure learning for discovering implicit traffic patterns from massive traffic data, causally-aware spatio-temporal multi-graph convolution network (CASTMGCN) for learning spatio-temporal dependencies, and conformal prediction for uncertainty quantification. In particular, CASTMGCN fuses several graphs that characterize different important aspects of traffic networks (including physical road structure, time-lagged causal effect, contemporaneous causal relationships) and an auxiliary graph that captures the effect of exogenous factors on the road network. On this basis, a conformal prediction approach tailored to spatio-temporal data is further developed for quantifying the uncertainty in node-wise traffic predictions over varying prediction horizons. Experimental results on two real-world traffic datasets of varying scale demonstrate that the proposed method outperforms several state-of-the-art models in prediction accuracy; moreover, it generates more efficient prediction regions than several other methods while strictly satisfying the statistical validity in coverage. An ablation study is also designed to examine the unique role of each considered graph in capturing traffic characteristics. This study not only contributes to the literature on traffic prediction\textemdash going beyond the conventional focus on connections of physical networks, but also provides a meaningful lens to support a wide range of learning problems with spatio-temporal characteristics.
}%




\maketitle
 
\section{Introduction}\label{sec:Intro}
Accurate and reliable prediction is a cornerstone in data-driven decision automation across a broad spectrum of applications, such as inventory control, traffic management, product recommendations, etc. To make accurate prediction, it is essential to discover the underlying mechanism governing the data generation process by learning from observational data data. The mechanism often gets manifested in various forms such as associations, causal relationships, and temporal/spatial dependencies, and should be encoded appropriately in the model under development\textemdash say, a machine learning (ML) model\textemdash to inform learning and prediction. Due to the variability nature of practical decision-making environments, prediction reliability is as important as, if not more important than, prediction accuracy. This necessitates a statistically meaningful estimation of prediction uncertainty, often in the form of a confidence interval that is anticipated to contain the ground truth at a given confidence level. 
Across a wide range of application domains, however, a rigorous framework to deliver high-quality prediction and statistically valid uncertainty estimation is largely lacking in the literature. In this paper, we take traffic prediction as an exemplary application to showcase a generic framework developed for making accurate and reliable prediction under uncertainty. The studied traffic prediction problem is featured by several traits prevalent in other domains (e.g., disease spread control, social network influence modeling, location-based recommendation service): discovery of implicit patterns from observed data, spatio-temporal dependencies learning, data-driven fusion of heterogeneous information, and uncertainty estimation. In this sense, the proposed approach is fairly generic as it can be readily extended to other spatio-temporal learning problems.

Traffic prediction is typically formulated as a spatio-temporal learning problem\textemdash that is, to use the traffic states in the past $M$ time steps to predict the states of the entire network over the next $H$ time steps. Due to complex temporal and spatial dependencies pervasive in traffic networks, making accurate and reliable traffic predictions is highly challenging. Luckily, deep neural networks have come to the rescue in tackling spatio-temporal forecasting problems due to its powerful ability in end-to-end learning of intricate spatio-temporal dependencies~\citep{shaygan2022traffic,li2022nonlinear}. 
The extant studies predominately characterize traffic patterns along the spatial and temporal dimensions separately. Temporal dependencies in spatio-temporal data are typically modeled using recurrent neural networks~\citep{luo2024gt}. Spatial dependencies, on the other hand, are captured with methods like convolution neural networks for grid data and graph convolutional networks (GCNs) for graph-structured data~\citep{pan2018spatial,li2022deep,li2018diffusion}. GCNs excel in extracting spatial features from non-Euclidean graph structures by relying on adjacency matrix defining node relationships built upon road topology, distances, similarities, and adaptive dynamic matrices~\citep{jiang2022graph}. However, most adjacency matrices only represent static spatial dependencies based on physical road network structure while overlooking dynamic implicit dependencies arising from causal drivers. These implicit dependencies propagate through the network over time and space and play an essential role in shaping the observed traffic patterns. A fixed adjacency matrix only reflects the physical connectivity of traffic network, but it fails to capture complex spatial traffic patterns. 

Various methods have been developed to remedy the shortcoming of static adjacency matrix by constructing additional matrices to model time-varying spatial dependencies, including Pearson correlation, Euclidean distance, and dynamic time warping~\citep{shaygan2022traffic}. However, these methods fall short in handling causality-related spatial dependencies as manifested in the implicit traffic patterns, although the importance of causality in traffic engineering has been well acknowledged in the literature~\citep{zhao2023causal}. For example, consider a network with two paths from origin $O$ to destination $D$: path $\text{p}_1 (r_1 \rightarrow r_3 \rightarrow r_4 \rightarrow r_6)$ and path $\text{p}_2 (r_1 \rightarrow r_2 \rightarrow r_5 \rightarrow r_6)$. If a driver learns about congestion on $r_5$ and takes route $\text{p}_2$, the shift in traffic flow increases the likelihood of congestion along road segment $r_3$. The congestion on $r_5$ spreads through the network over time, resulting in a likely congestion on $r_3$ eventually. A static adjacency matrix fails to model this implicit pattern because there is no direct link between $r_5$ (cause) and $r_3$ (effect). Unlike explicit patterns, it is challenging to model implicit traffic patterns and the resulting spatial dependencies due to the uncertainty in the propagation of causal effects through the traffic network over time and space.

To capture implicit patterns in complex systems, causal discovery has been extensively  applied across a broad spectrum of fields, including medicine, economics, and social sciences~\citep{glymour2019review,li2024fusion,feuerriegel2024causal}. In the context of traffic prediction, causal discovery also holds a great promise in informing interpretable feature engineering for downstream prediction and decision analytics tasks~\citep{du2023spatial}. Though discovering causal relationships in traffic network helps to understand implicit traffic patterns, mining causal relationships from massive multivariate data poses a significant computational challenge due to the combinatorial nature of causal discovery~\citep{glymour2019review}. As of now, only a limited number of studies have ever attempted to encode causal relationships into deep learning models for traffic-related tasks~\citep{yang2021dynamic,zhao2022stcgat,liang2023dynamic,luan2022traffic}. Analyzing causal structure in large-scale urban traffic networks is still in its infancy, as extant studies primarily focus on small-scale road networks with network size less than 100 nodes due to the high computational complexity. Moreover, few studies have considered explicit and implicit traffic patterns simultaneously, despite the critical role of causal graph in informing accurate traffic prediction.

In addition, quantifying the uncertainty in traffic predictions is essential for developing robust and resilient traffic management measures. Yet extant studies focus primarily on deterministic predictions, neglecting the inherent uncertainty in model predictions. To address this gap, recent research has explored various methods to estimate the uncertainty in spatio-temporal traffic prediction models. For instance, variable inference and deep ensemble methods have been employed to estimate aleatory and epistemic uncertainties~\citep{qian2023towards, qian2023uncertainty}. These approaches, however, often assume that traffic variables follow a fixed type of distribution (e.g., Gaussian) while the assumed probability distribution may not accurately capture the versatile nature of traffic data~\citep{jin2024spatial}. In fact, some existing studies~\citep[see, e.g.,][]{larsson2014design} find that traffic states may follow a negative Binomial or Poisson distribution rather than Gaussian distribution. To remedy this,~\citet{wu2023adaptive} used a distribution-free uncertainty quantification approach\textemdash conformal prediction\textemdash to generate prediction region at a specified confidence level. Importantly, the generated prediction region is valid in distribution-free scenarios as they possess explicit, non-asymptotic guarantee without assumptions on probability distribution~\citep{balasubramanian2014conformal}. However, conformal prediction (CP) is built upon data exchangeability while time-series traffic data exhibits strong temporal dependency. As a result, there is no provable statistical guarantee to ensure that the uncertainty quantified in the context of spatial-temporal data modeling is statistically valid, and the biased uncertainty estimation might lead to suboptimal decisions in traffic system optimization.

To sum up, several long-standing challenges in traffic prediction remain to be tackled. First, extant literature focuses primarily on modeling explicit traffic patterns arising from the road network topology, but does not account for causality-related implicit traffic patterns that are essential for accurate network-wide traffic prediction. Second, only a limited number of studies attempt to quantify the uncertainty in traffic predictions, and most of them simplify the probability distributions used for traffic modeling. As a consequence, they might fail to capture versatile traffic conditions in practice. 
Third, existing methods for uncertainty estimation provide no provably formal guarantee on the statistical validity of confidence interval built upon the estimated uncertainty. The confidence interval is likely to deviate from decision maker's expectation on its chance of containing the ground truth value. In addition, existing traffic prediction models are often tested upon data collected from a small or local region, like the METR-LA traffic dataset containing 207 sensors from Los Angeles, USA~\citep{jagadish2014big}. Modeling spatio-temporal characteristics of large-scale urban traffic networks enables a nuanced understanding of city-wide traffic phenomena and a more comprehensive appraisal of traffic prediction models. To fill these gaps, we develop a causally-aware traffic prediction model capable of characterizing temporal and explicit/implicit spatial dependencies to generate deterministic predictions. The causality-related spatial dependency is discovered from massive traffic data and then seamlessly encoded in the developed deep learning model while the effect of difficult-to-model exogenous factors (e.g., population density, road conditions) is captured by an auxiliary graph. Given the deterministic model, we further develop a conformal prediction method dedicated to spatio-temporal data for estimating the prediction uncertainty by generating a statistically valid and efficient prediction region (also called confidence interval in statistics) for each node. Compared to the state-of-the-art literature, our work makes the following contributions: 

\begin{enumerate}
\item  We leverage DYNOTEARS to discover implicit traffic patterns in the form of time-lagged and contemporaneous causal relationships by learning from massive historical traffic data. The discovered causal graphs along with the explicit traffic pattern manifested as adjacency matrix are seamlessly encoded into deep learning model to inform accurate traffic prediction. 


\item We develop a distribution-free conformal prediction method to quantify the uncertainty in the network-wide traffic prediction. The proposed conformal prediction approach yields efficient and statistically valid prediction regions at the network level in the finite sample case.

\item We collect and prepare a large-scale traffic flow dataset\textemdash the Hong Kong dataset having 607 nodes and 4-month of traffic flow data. The dataset covers major roads throughout Hong Kong and serves as a useful large-scale testbed to benchmark the performance of traffic prediction models.
	
\item We perform extensive experiments on two real-world traffic datasets to validate the model performance. The developed model outperforms several state-of-the-art models in terms of prediction accuracy. Besides, the proposed conformal prediction method not only generates statistically valid prediction regions, but also produces more efficient prediction regions than other methods. 
\end{enumerate}

The rest of this paper is structured as follows: Section~\ref{sec:literature_review} offers a brief review on the relevant literature. Section~\ref{sec:Methodology} outlines the traffic prediction problem and details our proposed approach. Section~\ref{sec:Computational Experiments} assesses the performance of our method using real-world traffic data from Hong Kong and Los Angeles, and compares it with several leading methods. Section~\ref{sec:Conclusion} summarizes our findings and suggests directions for future research. Additional results are available in the Online Supplement.

\vspace{-0.5em}
\section{Literature Review }\label{sec:literature_review}
\subsection{Traffic prediction}
Existing approaches for traffic prediction can be categorized into statistical methods, machine learning methods, and deep learning methods. Early research focused on statistical  methods, but they struggle with capturing complex traffic patterns and long-term temporal dependencies in traffic data. Machine learning models, such as support vector regression, have been used to model complex traffic patterns, but they require dedicated feature engineering to manually design and extract useful features from raw traffic data~\citep{shaygan2022traffic}. Recently, deep learning models like recurrent neural networks and temporal convolutional networks (TCNs) have been exploited to automatically extract features from time-series data for traffic-related tasks, such as network traffic prediction and sequential route recommendation~\citep{shaygan2022traffic, liu2023cost}. Apart from temporal dependency, accurate traffic prediction also necessitates sound modeling of spatial dependency in road network. Road networks can be represented either as a regular grid structure or a non-Euclidean topology. In spatial dependency modeling, convolutional neural networks are commonly used for processing grid structures to extract spatial features~\citep{pan2018spatial}, but grid-based representation fails to capture the interior spatial inter-connectivity among roads. Non-Euclidean topology often represents road networks as graphs with nodes denoting in-roadway sensors or intersections. GCNs excel at modeling spatial dependencies among nodes~\citep{kipf2016semi, shaygan2022traffic} via performing graph convolutions over adjacency matrix of the graph to extract abstracted spatial representations on road networks. Several models established upon adjacency matrix have been developed to capture spatial dependencies in graphs, including diffusion convolutional recurrent neural networks~\citep{li2018diffusion}, and spatio-temporal GCNs~\citep{yu2017spatio}. As each type of adjacency matrix models a specific aspect of spatial relationships, combining multiple matrices facilitates characterizing multifaceted spatial relationships. For instance,~\citet{xiu2022modelling} created distance, attraction, and pattern-related adjacency matrices to forecast network-wide metro system passenger flow. Considering hidden spatial dependencies,~\citet{wu2019graph} introduced a self-adaptive adjacency matrix as learnable parameters for allowing the model to discover hidden spatial dependencies from traffic data, leading to a superior performance in various traffic prediction tasks.

\vspace{-0.5em}
\subsection{Dynamic Causal Structure Learning}
When learning causal structures from observational data, Bayesian networks (BNs) are a popular analytical framework for modeling probabilistic relationships among random variables. Typically, BNs use directed acyclic graphs (DAGs) to represent conditional dependencies among random variables, where nodes denote random variables and directed edges indicate conditional dependencies~\citep{heckerman2008tutorial,lan2024multifun}. Dynamic BNs extend static BNs to handle time-series data by modeling temporal dependencies~\citep{murphy2002dynamic}. Currently, there are two classes of methods for learning graph structures: score-based and constraint-based methods~\citep{cheng2001learning}. Score-based methods aim to identify an optimal graph structure that maximizes a preset scoring function, such as the Bayesian Information Criterion or Akaike Information Criterion, by searching over a large space of possible network structures~\citep{de2011efficient}. Constraint-based methods use conditional independence tests to identify causal structures by iteratively performing statistical tests to determine which variables are conditionally independent of each other given the rest of the variables~\citep{behjati2018order}. Score-based methods are often computationally intensive as the search space grows exponentially with the number of observed variables. Recent advances have emphasized on improving the computational efficiency of causal structure learning. For example,~\citet{zheng2018dags} proposed NOTEARS to transform discrete combinatorial optimization into a continuous problem for efficient discovery of large-scale causal structure. Building on this,~\citet{pamfil2020dynotears} developed DYNOTEARS--a dynamic version of NOTEARS--to discover temporal dependencies, thus offering a scalable approach for learning causal structure from time-series data. In  traffic domain, causal inference also plays a crucial role for understanding complex traffic phenomena~\citep{yang2021dynamic,liu2024estimating}. For example,~\citet{luan2022traffic} integrated Bayesian inference into a deep learning framework to build a dynamic Bayesian GCN and established a set of dynamic propagation rules to understand congestion in road networks from historical data. 

\vspace{-0.5em}
\subsection{Uncertainty Quantification}
Traffic systems are inherently complex and dynamic, influenced by various factors such as road conditions, weather. Day-to-day variability, fluctuations in travel demand, and random traffic incidents all contribute to the uncertainty in traffic prediction~\citep{sengupta2024bayesian}. Quantifying the uncertainty arising from these sources enables the development of robust control and congestion mitigation measures. Bayesian methods have been widely used for quantifying prediction uncertainty in ML models~\citep{sengupta2024bayesian,lin2024deterring} by assuming that the parameters of ML models follow certain prior distributions. These prior distributions are then combined with the observed data to infer posterior distributions of model parameters via Bayesian inference.~\citet{jin2024spatial} assumed traffic flow follows a Gaussian distribution and added an additional layer to capture uncertainty in traffic state evolution.~\citet{qian2023towards} developed two independent subnetworks to maximize heterogeneous log-likelihood for estimating aleatoric uncertainty in traffic flows and utilized multi-hierarchical conformal calibration to relax the Gaussian distribution assumption. Other studies, such as~\citet{wang2024uncertainty}, and \citet{jiang2023uncertainty}, have measured uncertainty by assuming variables follow either a Gaussian or Tweedie distribution. Despite these advancements, the inherent variability in traffic variables makes it challenging to characterize their uncertainty with a fixed probability distribution. Traffic variables can exhibit diverse characteristics over time due to numerous internal and external factors~\citep{hall1996traffic}, such as driving behavior, population density, weather conditions, and rush hour dynamics. Moreover, existing methods for uncertainty estimation often lack a formal guarantee on the statistical validity of the confidence intervals derived from estimated parameters. Addressing these challenges requires developing a generic method to provide robust statistical guarantees for supporting reliable decision-making in traffic management.

Table~\ref{tab:overview} provides a side-by-side comparison of our work against existing traffic prediction studies. To our knowledge, our approach of adaptively fusing multiple individual graphs—each representing a specific aspect of traffic networks—within GCNs is unprecedented. Moreover, our approach is the first of its kind in the literature to estimate uncertainty associated with spatio-temporal traffic data in a distribution-free and statistically valid manner. 

\vspace{-0.5em}
\section{Methodology}\label{sec:Methodology}
In this section, we describe the proposed methodology for network-wide traffic prediction. We first define the traffic prediction problem, formulate it as a learning problem, and outline the learning objective (Section~\ref{sec:Problem Statement}). Next, we introduce the \underline{C}ausally-\underline{A}ware \underline{S}patio-\underline{T}emporal \underline{M}ulti-\underline{G}raph \underline{C}onvolutional \underline{N}etwork (CASTMGCN) for making deterministic prediction for each node (Section~\ref{sec:models}). Finally, we explain how to use the conformal prediction method to generate statistically meaningful uncertainty estimations for spatio-temporal traffic data (Section~\ref{sec:uq}). The mathematical notations used in this paper are summarized in Table~\ref{tab:notation} (see the Online Supplement).

\vspace{-0.5em}
\subsection{Problem Description}\label{sec:Problem Statement}
Traffic prediction involves two key elements: traffic network and traffic state, described as below.

\textbf{Traffic network.} A traffic network can be mathematically represented as a graph $G = (V, E, \bm{A})$, where $V$ is a finite set of nodes $(|V|=N) $ representing sensors, roads, intersections, or stations. $E$ is the set of edges indicating connectivity between nodes, and $\bm{A} \in \mathbb{R}^{ N \times N} $ is the adjacency matrix of graph $G$. In this study, we treat each road segment as an individual node in graph $G$ since traffic detectors are installed underground to record relevant traffic quantities (e.g., traffic volume, speed). 

\textbf{Traffic state.} The traffic variables of interest at time $t $ are denoted by $ \bm{x}_t \in \mathbb{R}^{N \times F}$, where $N$ is the number of nodes in the road network and $F$ is the number of traffic state variables (e.g., volume, speed, density). Let $\bm{X}_M =[\bm{x}_{t-(M-1)}, \bm{x}_{t-M}, \ldots, \bm{x}_t] \in \mathbb{R}^{N \times F \times M}$ represent the traffic variables of the entire road network over the $M$ time steps prior to (and including) $t$.



\begin{figure}[!ht]
	\centering
	\includegraphics[scale=0.372]{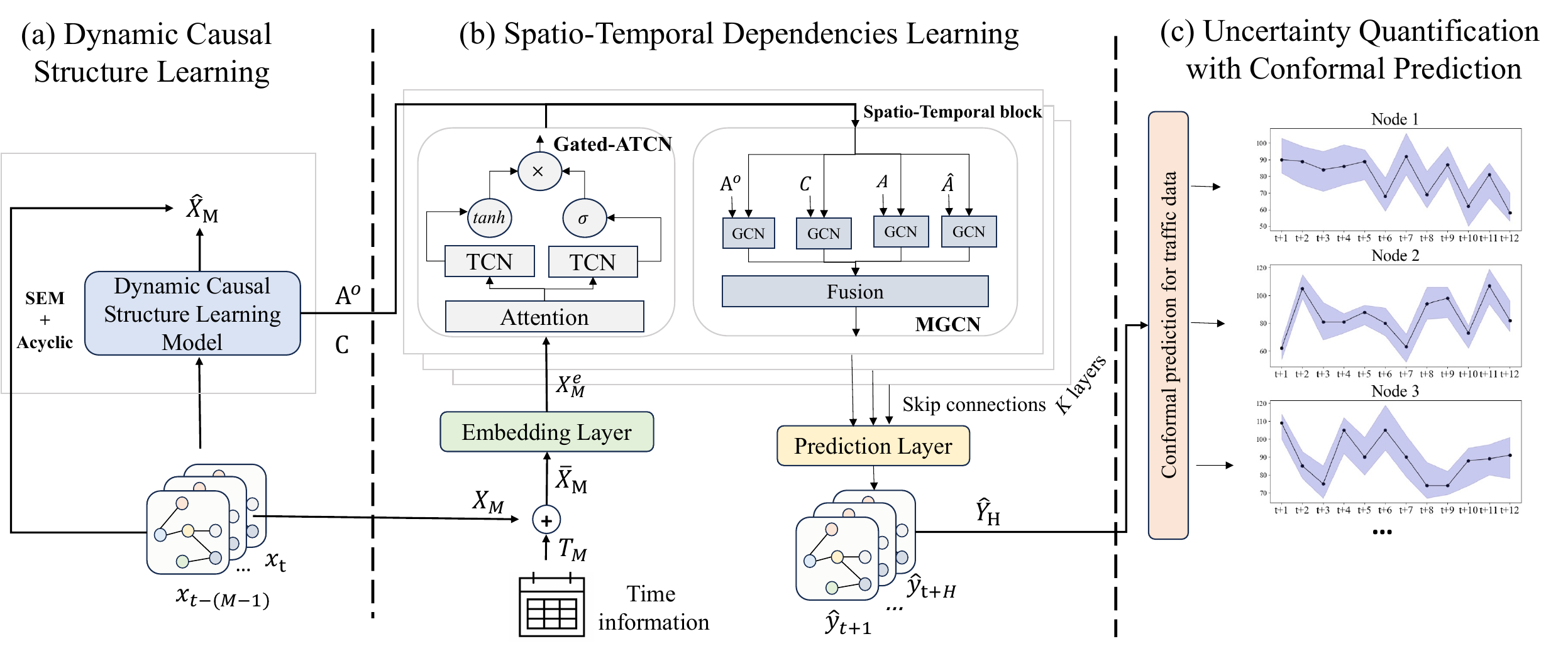}
	\caption{Flowchart of the Proposed CASTMGCN Methodology}
	\label{fig:4}
\end{figure}

Given graph $G$ and historical traffic states $\bm{X}_M$, the traffic prediction problem is mathematically formulated as learning a function $f(\cdot)$ to predict the traffic state of the $N$ nodes over the next $H$ steps (i.e., $\hat{\bm{Y}}_H=\left[\hat{\bm{y}}_{t+1}, \hat{\bm{y}}_{t+2}, \ldots, \hat{\bm{y}}_{t+H}\right] \in \mathbb{R}^{N \times F \times H}$) based on the historical traffic states $\bm{X}_M$ and the road network $G$. In this work, we devise CASTMGCN for generating accurate and reliable network-wide traffic prediction. The methodology involves three main steps (Figure~\ref{fig:4}): dynamic causal structure learning, spatio-temporal dependencies learning, and uncertainty quantification with conformal prediction. First, historical traffic data is used to discover time-lagged and contemporaneous causal relationships among nodes, creating causal graphs to model implicit traffic patterns. Next, spatio-temporal dependencies are captured using a predefined adjacency matrix, a pretrained causal matrix, and an auxiliary adaptive matrix to account for the impact of exogenous factors like population density and road characteristics. The prediction layer produces deterministic point forecast for future traffic states. Finally, conformal prediction for spatio-temporal data (CPST) quantifies the uncertainty of each deterministic prediction, providing a reliable prediction interval based on a specified coverage target. Next, we elaborate on each component of the proposed methodology.

\vspace{-1.3em}
\vspace{-2em}

\subsection{Dynamic Causal Structure Learning}\label{sec:models}
\vspace{-1.4em}
\begin{figure}[!ht]
	\centering
	\includegraphics[width=0.65\linewidth]{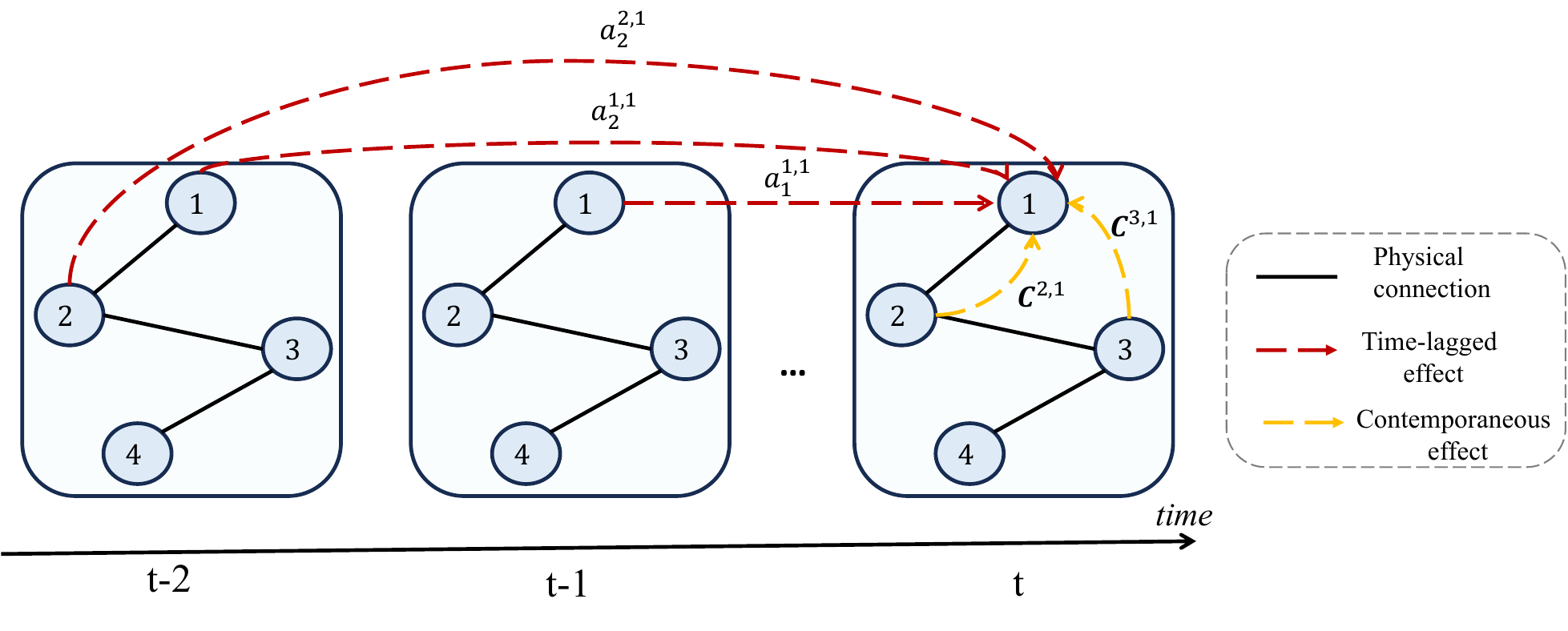}
    \caption{Demonstration of Intra-slice and Inter-slice Causal Dependencies in Traffic Networks. Note: The network has 4 nodes and the time-lagged value is set as 2 in this example. (Only draw the lines that have an effect on $\bm{x}_t^{(1)}$, i.e., the traffic states of node $1$ at time $t$.) }
	\label{fig:2}
\end{figure}
\vspace{-1.8em}

As previously mentioned, many implicit traffic patterns can be modeled from the lens of causality. From a causal perspective, the traffic state at time $t$ (i.e., $\bm{x}_t$) is influenced by two major factors: time-lagged effects (inter-slice temporal dependencies) and contemporaneous effects (intra-slice dependencies). Consider the traffic state of node 1 at time $t$, as illustrated in Figure~\ref{fig:2}. Node 2 does not affect node 1 until time $t$ due to the delay in the propagation and accumulation of causal effects over time and space. We represent the time-lagged causal effect of $\bm{x}_{t-2}$ on $\bm{x}_t$ using a squared matrix $\bm{A}_{2}^{\circ}$ of size $N \times N$, where entry $a_{2}^{2,1}$ in this matrix characterizes the time-lagged effect of node 2 at time $t-2$ on the traffic state of node 1 at time $t$. Moreover, it is crucial to consider contemporaneous causal effects depicting the dependencies among traffic state variables measured at the same time. As shown in Figure~\ref{fig:2}, we use another matrix $\bm{C}$ of size $N \times N$ to quantify the contemporaneous effects. For example, $\bm{C}^{3,1}$ indicates the contemporaneous effect of node 3 on node 1 at time $t$.

To quantify time-lagged causal effects, we consider a maximum number of time steps $P$, known as the autoregressive order ($P \leq M$). Assuming a stable data generation process over time~\citep{pamfil2020dynotears} and that causal effects are linear, the observed time series can be described using a structural vector autoregressive model~\citep{swanson1997impulse}:
\begin{equation}\label{equ2}
\bm{x}_t^\texttt{T}=\bm{x}_t^\texttt{T} \bm{C}+\bm{x}_{t-1}^\texttt{T} \bm{A}_{1}^{\circ}+\cdots+\bm{x}_{t-P}^\texttt{T} \bm{A}_{P}^{\circ}+\bm{Z}_{t}^\texttt{T},
\end{equation}
where $\bm{C} \in \mathbb{R}^{N \times N}$ indicates the intra-slice causal dependency, $\bm{A}_i^{\circ} \in \mathbb{R}^{N \times N}$ $(i \in \{1,…,P\})$ quantifies the inter-slice dependency of the traffic state $\bm{x}_{t-i}$ on the current traffic state $\bm{x}_{t}$, and $\bm{Z}_{t} \in \mathbb{R}^{N \times F}$ is a centered error variable to account for exogenous variables the model does not consider ($\bm{Z}_{t}$ is independent within and across time). Matrices $\bm{C}$ and $\bm{A}_i^{\circ}$ $(i \in \{1,2,…, P\})$ are weighted causal matrices with non-zero entries representing the contemporaneous and time-lagged causal effects. 

The objective of causal structure discovery is to learn matrices $\bm{C}$ and $\bm{A}_i^{\circ}$ $(i \in \{1,2,…,P\})$ from historical traffic data. As causal effect is modeled in the form of DAG, no cycle is permitted in the learnt graph. Since $\bm{A}_i^{\circ}$ $(i \in \{1,2,…,P\})$ represents the time-lagged causal effect, the edges in $\bm{A}_i^{\circ}$ are only allowed to go forward in the time space and thus no cycle exists naturally. As a result, the acyclicity constraint narrows down to guarantee that the learnt matrix $\bm{C}$ is acyclic. To this end, we leverage the DYNOTEARS developed by~\citet{pamfil2020dynotears} to learn causal graph from massive historical traffic data (see Section~\ref{sec:Computational Experiments} for description on the traffic dataset). The discrete DAG constraint on $\bm{C}$ is replaced with a continuous and smooth equality constraint ${h(\bm{C})} = 0$. Mathematically, ${h(\bm{C})}$ is defined as $h(\bm{C}) = \text{tr}(e^{\bm{C} \odot \bm{C}} - N)$, where $\text{tr}(\cdot)$ is the trace of a matrix, $\odot$ denotes element-wise product of two matrices, and $N$ is the dimension of matrix $\bm{C}$. In essence, $h(\cdot)$ measures the DAGness of a matrix.~\citet{zheng2018dags} demonstrated that $h(\bm{C}) = 0$ is satisfied if and only if $\bm{C}$ is acyclic, providing a smooth and exact function to encode the acyclicity constraint.

Let $\bm{S} = [\bm{x}_{t-1}^\texttt{T}, \bm{x}_{t-2}^\texttt{T}, \cdots, \bm{x}_{t-P}^\texttt{T}]$ and ${\bm{A}^{\circ}} = [ {\bm{A}_1^{\circ},\;\bm{A}_2^{\circ}, \cdots, \bm{A}_P^{\circ}}]^\texttt{T}$. Built upon the continuous acyclicity constraint, causal structure learning is formulated as a constrained optimization problem as below:
\begin{equation}\label{eq:constrained_optimization_learning}
\begin{aligned}
\textstyle  \min_{\bm{C},\bm{A}^{\circ}}\;\ & l(\bm{C},\bm{A}^{\circ}) = \frac{1}{2n} \lVert \bm{x}_t^\texttt{T}-\bm{x}_t^\texttt{T}\bm{C}-\bm{S}\bm{A}^{\circ} \rVert ^2_F + \lambda_{\bm{C}} {\left\| \bm{C} \right\|_1} + \lambda_{\bm{A}^{\circ}} {\left\| \bm{A}^{\circ} \right\|_1}, \;\;
    \text{s.t. } & h(\bm{C})=0,
\end{aligned}    
\end{equation}
where $n$ is the size of training data, $\lVert \cdot \rVert^2_F$ is the squared Frobenius norm, $\lVert \cdot  \rVert_1$ is the $L_1$ norm, and $\lambda_{\bm{C}}$ and $\lambda_{\bm{A}^{\circ}}$ indicate the weights associated with the regularization terms for $\bm{C}$ and $\bm{A}^{\circ}$, respectively. For the sake of brevity, the noise term $\bm{Z}_t^\texttt{T}$ is omitted. The loss function $l(\bm{C},\bm{A}^{\circ})$ consists of two parts: The first part (i.e., $\frac{1}{2n} \lVert \bm{x}_t^\texttt{T}-\bm{x}_t^\texttt{T}\bm{C}-\bm{S}\bm{A}^{\circ} \rVert ^2_F$) is a loss term to measure the fitness of a learnt model to the observed data. The second part (i.e., $\lambda_{\bm{C}} {\| \bm{C} \|_1} + \lambda_{\bm{A}^{\circ}} {\| \bm{A}^{\circ}\|_1}$) regularizes the learnable parameters to induce sparsity in the learnt causal matrices $\bm{C}$ and $\bm{A}^{\circ}$, in order to avoid model overfitting. 



By recasting the discrete optimization problem as a continuous one, standard numerical optimization algorithms, such as Limited-memory Broyden–Fletcher–Goldfarb–Shanno algorithm~\citep{nocedal1999numerical}, can be used to solve the causal structure learning problem. As shown in Figure~\ref{fig:4}, the learned causal relationships in $\bm{C}$ and $\bm{A}^{\circ}$ are then fed into a spatio-temporal dependencies learning module to enable causally-aware traffic prediction over the entire road network.

\vspace{-0.5em}
\subsection{Spatio-Temporal Dependencies Learning}\label{sec:multi_graph_fusion}
The matrices $\bm{C}$ and $\bm{A}^{\circ}$ derived from causal structure learning along with the predefined adjacency matrix indicating physical connectivity are incorporated into neural network for traffic prediction. Figure~\ref{fig:4} shows the neural network composed of an embedding layer, stacked spatio-temporal blocks (ST-blocks), and a prediction layer. Each ST-block combines a gated attention temporal convolution network (Gated-ATCN) for temporal dependencies and a multi-graph convolution network (MGCN) for spatial dependencies in traffic data. By stacking these blocks, the module effectively captures spatio-temporal features at various scales, thus enhancing the accuracy of traffic prediction.

\vspace{-0.5em}
\subsubsection{Embedding Layer.}
When making traffic prediction, it is crucial to encode specific time information (e.g., peak vs non-peak hours, weekdays vs weekends) into neural networks to capture time-dependent traffic patterns and causal relationships to enhance prediction accuracy~\citep[e.g.,][]{zheng2020gman,guo2021learning,wu2019graph}. Herein, we incorporate essential time information at granularity of 5 minutes (e.g., dividing an hour into 12 time steps), the hour of the day (e.g., 1, 2, 3, $\cdots$), and the day of the week (e.g., Monday, Tuesday, $\cdots$). Let $\bm{T}_M$ denote the one-hot representation of time-related information over $M$ time steps. We concatenate $\bm{X}_M$ with $\bm{T}_M$, forming the extended traffic state $\overline{\bm{X}}_M = \bm{X}_M || \bm{T}_M \in \mathbb{R}^{N \times M \times F'}$, where $F'$ is the feature space dimension. Additionally, an embedding layer maps the original node-specific data across time steps into a $d$-dimensional continuous representation to enhance learning effectiveness. The embedding operation is represented as $\bm{X}_M^e = p(\overline{\bm{X}}_M,\bm{W}_e)$,
where $\bm{X}_M^e = [\bm{X}_M^{(1)}, \bm{X}_M^{(2)}, \cdots, \bm{X}_M^{(N)}] \in \mathbb{R}^{N \times M \times d}$ denotes the extended traffic state after embedding, $\bm{W}_e$ denotes the learnable parameters, and $p: \overline{\bm{X}}_M \rightarrow \bm{X}_M^e$ is the embedding function. After embedding, the first two dimensions of $\overline{\bm{X}}_M$ remain unchanged, while the last dimension changes from $F'$ to $d$ in mapping discrete variables to continuous representations.

\vspace{-0.5em}
\subsubsection{Gated-ATCN.}
While recurrent neural network are commonly used to capture temporal dependencies, they exhibit limitations when processing long temporal sequences, such as shorter reference windows, diminishing signals from earlier data, and exploding/vanishing gradients during backpropagation~\citep{lu2020spatiotemporal}. To address these issues, we develop a Gated-ATCN to capture both short- and long-term temporal dependencies in time-series traffic data. We leverage the gated mechanism of TCNs for handling long sequences~\citep{wu2019graph} and the self-attention mechanism for selectively capturing important information in the time-series data~\citep{vaswani2017attention}.	

\vspace{-1.5em}
\begin{figure}[!ht]
	\centering
	\includegraphics[scale=0.29]{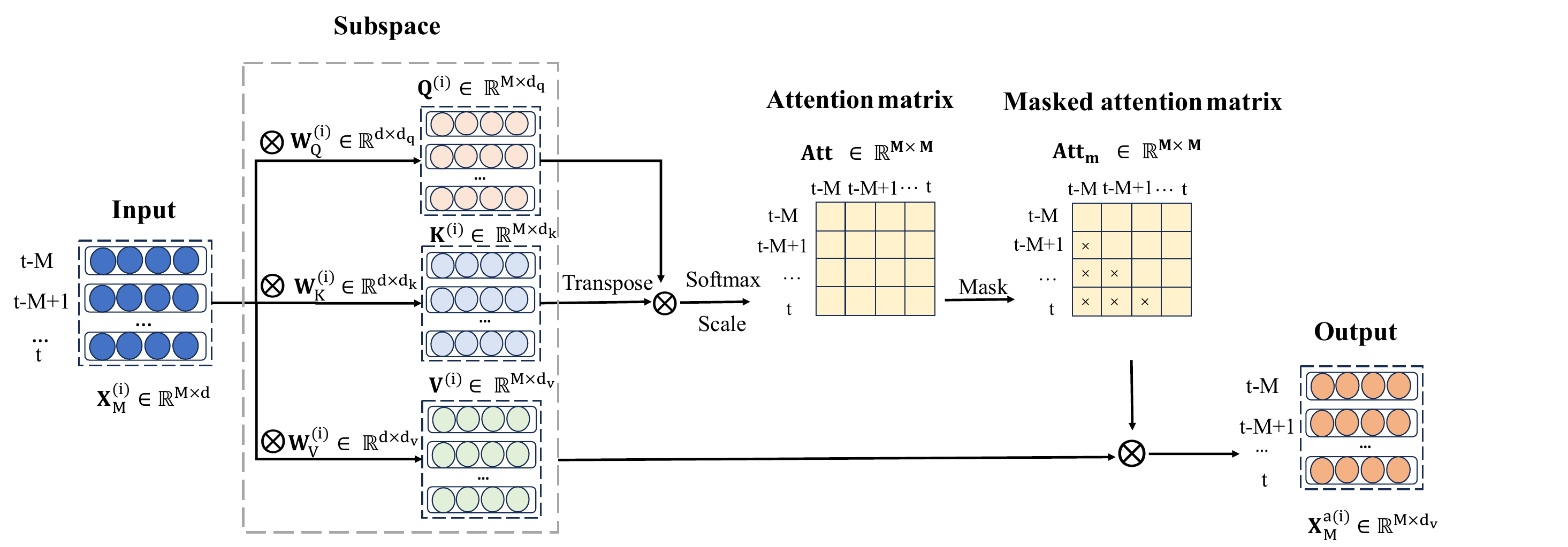}
	\caption{Illustration of Self-Attention Mechanism. The traffic state of node $i$ is used for demonstration purpose}
	\label{fig:Attention}
\end{figure}
\vspace{-1.6em}

In the Gated-ATCN, the self-attention mechanism assigns weights to traffic states at preceding time steps (e.g., $t- M$, $t- M+1$, $\cdots$, $t-1$) in the input sequence to dynamically adjust their impact on the traffic state at time $t$. This allows the model to focus on the most relevant inputs when making traffic prediction. The basic procedure of the self-attention mechanism is illustrated in Figure~\ref{fig:Attention}. Given the representation $\bm{X}_M^{(i)} \in \mathbb{R}^{M \times d}$ of node $i$ in the embedding space, the self-attention mechanism transforms $\bm{X}_M^{(i)}$ into three quantities: query $\bm{Q}^{(i)} \in \mathbb{R}^{M \times d_q}$, key $\bm{K}^{(i)} \in \mathbb{R}^{M \times d_k}$, and value $\bm{V}^{(i)} \in \mathbb{R}^{M \times d_v}$. This is done via linear transformation with three weight matrices $\bm{W}^{(i)}_Q \in \mathbb{R}^{d\times d_q}$, $\bm{W}^{(i)}_K \in \mathbb{R}^{d\times d_k}$, and $\bm{W}^{(i)}_V \in \mathbb{R}^{d\times d_v}$, respectively. To be specific, $\bm{Q}^{(i)} = \bm{X}_M^{(i)} \bm{W}^{(i)}_Q$, $\bm{K}^{(i)} = \bm{X}_M^{(i)} \bm{W}^{(i)}_K$, and $\bm{V}^{(i)} = \bm{X}_M^{(i)} \bm{W}^{(i)}_V$. The query and key vectors derive the attention score to indicate the focus for each time step. The value vector creates a weighted combination of the attention score and the transformed traffic state. To ensure that predictions rely only on past time steps, future traffic information is masked by setting entries in a mask matrix $\bm{M}$ to $-\infty$. All entries in the lower triangular part of $\bm{M}$ (except the diagonal) are $-\infty$. With the masked attention matrix, we derive the attention score as:
\begin{equation}
\textstyle \bm{X}_M^a =  \text{softmax}\left(\frac{\bm{Q}\bm{K}^T + \bm{M}}{\sqrt{d_k}}\right)\bm{V},
\end{equation}
where $\text{softmax}\left(\frac{\bm{Q}\bm{K}^T + \bm{M}}{\sqrt{d_k}}\right)$ corresponds to the attention matrix $\text{Att}_m$ in Figure~\ref{fig:Attention}, $d_k$ is the dimension of the query vector. Applying softmax ensures that the derived attention scores are positive and sum to one. The attention scores form a distribution over the input sequence representing the relative importance of traffic states in preceding time steps for encoding the traffic state at time $t$. In essence, self-attention encodes the impact of preceding traffic states on the current state for each node via the learned attention matrix $\text{Att}_m$. Temporal dependency is translated into the self-attention output $\bm{X}_M^{a(i)} \in \mathbb{R}^{M \times d_v}$ for each node $i$ (with $\bm{X}_M^a=[\bm{X}_M^{a(1)}, \bm{X}_M^{a(2)}, \cdots, \bm{X}_M^{a(N)}]$).


Next, we feed $\bm{X}_M^{a(i)}$ into gated-TCN and use convolution kernel filters to learn useful representations for making traffic prediction. 
TCN performs convolutions over the time domain, where each neuron receives input from a restricted area of the previous layer (neuron's receptive field). Unlike standard CNNs, TCN uses dilated convolutions to expand the receptive field to enhance temporal dependency modeling~\citep{bai2018empirical}. In this study, the self-attention output $\bm{X}_M^{a(i)}$ undergoes three dilated convolutions with dilation rates of 1, 2, and 4. Hyperparameters of convolution kernel filter (i.e. filter size, padding)  are tuned with the validation data. 
To enhance the model's ability to capture complex temporal dependency, we introduce a gated mechanism to filter out less informative features with the sigmoid function $\sigma(\cdot)$ and scale features to the range $[-1, 1]$ using the hyperbolic tangent function $\text{tanh}(\cdot)$ as well. For node $i$, the gated mechanism operates as follows:
\begin{equation}
\textstyle \bm{X}_c^{(i)} = \tanh(\bm{W}_g\star \bm{X}_M^{a(i)}+\bm{b}_g) \odot \sigma (\bm{W}_{\sigma} \star \bm{X}_M^{a(i)} +\bm{b}_\sigma),
\end{equation}
where $\star$ indicates the dilated convolution operation, $\bm{W}_g$ and $\bm{W}_\alpha$ are kernel filters of the dilated convolution, and $\bm{b}_g$ and $\bm{b}_{\alpha}$ are biases. Note that all of $\bm{W}_g$, $\bm{W}_\alpha$, $\bm{b}_g$, and $\bm{b}_{\alpha}$ are trainable parameters.

\vspace{-0.4em}
\subsubsection{MGCN.}

To extract spatial dependencies from traffic data, we develop an MGCN to learn  spatial relationships in the road network. To compensate the effect of overlooked exogenous factors, we introduce an auxiliary matrix $\hat{\bm{A}}$ to adaptively characterize their spatial impact on the traffic state. Next, MGCN fuses multiple individual graphs representing different spatial relationships to extract useful spatial features~\citep{geng2019spatiotemporal}. At a high level, MGCN is formulated as:
\begin{equation}\label{eq:MGCN}
\textstyle \bm{X}_s = \bigsqcup_{a \in \mathbf{A}_{\text{all}}} \sum_{k=0}^{K} g(a)^k \bm{X}_c \bm{W}_k,
\end{equation}
where $\bm{X}_s$ indicates the learnt spatially-aware feature representation, $g(a)$\textemdash indicates the graph processing operator used for processing each type of graph (see~\ref{sec:g_a} in the online supplement for more details), $\bm{W}_k$ is a trainable weight matrix to transform the extracted spatially-aware feature representation into the corresponding traffic state, $\bigsqcup$ denotes the data fusion operation (e.g., sum, max, mean, weighted average), and $\mathbf{A}_{\text{all}}$ represents the set of graphs considered by MGCN, including the static adjacency matrix ($\bm{A}$), the learnt causal matrices ($\bm{A}^{\circ}, \bm{C}$), and the auxiliary matrix ($\hat{\bm{A}}$).

As mentioned earlier, we introduce an auxiliary matrix $\hat{\bm{A}}$ to adaptively learn the spatial dependency overlooked by $\bm{A}$, $\bm{A}^{\circ}$, and $\bm{C}$. Following~\cite{wu2019graph}, we randomly initialize two learnable node embeddings $\hat{\bm{A}}_1 \in \mathbb{R}^{N\times d_n}$ and $\hat{\bm{A}}_2 \in \mathbb{R}^{N \times d_n}$ and update their values by
\begin{equation}\label{eq:A_hat}
\textstyle \hat{\bm{A}} = \text{softmax}(\text{ReLU}(\hat{\bm{A}}_1\hat{\bm{A}}_2^T)).
\end{equation}

Note that both $\hat{\bm{A}}_1$ and $\hat{\bm{A}}_2$ are learnable parameters. MGCN uses customized graph operators to capture spatial dependencies: adjacency matrix ($\bm{A}$) for physical road network structure, causal graphs ($\bm{A}^{\circ}$, $\bm{C}$) for capturing implicit traffic patterns, and auxiliary matrix ($\hat{\bm{A}}$) for modeling spatial relationships arising from exogenous factors. Combined together, Gated-ATCN and MGCN form an ST-block to learn spatio-temporal dependencies autonomously.

\vspace{-0.6em}
\subsubsection{Prediction Layer.}
Following \citet{he2016deep}, we add residual connections to each component (i.e., Gated-ATCN and MGCN) to prevent gradient vanishing during backpropagation. Stacking multiple ST-blocks enhances the model's capacity to learn spatio-temporal dependencies. Assuming there are $k$ ST-blocks, the output of the final ST-block is represented as $\bm{\hat{Y}} = \bm{X}_s^k + \bm{X}_{\text{s}}^{k-1}$, where $\bm{\hat{Y}} = [\bm{\hat{Y}}_{t+1},\bm{\hat{Y}}_{t+2},...,\bm{\hat{Y}}_{t+H}]$ is the final output of the model, $\bm{X}_s^k$ is the output produced by the $k$-th ST-block, and $\bm{X}_{\text{s}}^{k-1}$ is the output from the $(k-1)$-th ST-block added to the $k$-th ST-block's output through the residual connection. 

Finally, deterministic traffic predictions over multiple look-ahead time steps are made after two nonlinear fully-connected layers. The model is trained to minimize the mean absolute errors between actual traffic observations and predictions over the road network using the loss function below:
\begin{equation}\label{eq:objective_function}
\textstyle	\mathcal{L} = \sum_{j = t+1}^{t+H} ||\bm{Y}_j - \hat{\bm{Y}}_j||,
\end{equation}
where $H$ defines the number of look-ahead time steps. To minimize the loss function defined in \eqref{eq:objective_function}, the model optimizes several learnable parameters through backpropagation. 

\subsection{Uncertainty Quantification with Conformal Prediction}\label{sec:uq}
Conformal prediction (CP) is a distribution-free method for uncertainty quantification. In regression problems, CP generates a prediction region $\Gamma^{1-\alpha}$ at a significance level $1-\alpha$ to ensure that $\Gamma^{1-\alpha}$ contains the ground truth value with at least a probability of $1-\alpha$ \citep{stankeviciute2021conformal}. The proposed CP method, following the general procedure of CP, splits the original training data into a training  $D_{\text{train}}$ and calibration set $D_{\text{calib}}$ with $n_{\text{train}} = |D_{\text{train}}|$ and $n_{\text{calib}}=| D_{\text{calib}}|$. A deterministic traffic prediction model $\hat{f}$ is first fitted on $D_{\text{train}}$. Next, a nonconformity score function $V(\cdot)$ is constructed to measure the nonconformity between the ground truth value and model prediction for each sample in $D_{\text{calib}}$. By employing absolute residual $r_i=V(\bm{X}_i,\bm{Y}_i)= |Y_i-\hat{Y}_i|, \; \forall \{(\bm{X}_i, \bm{Y}_i)\} \in D_{\text{calib}}$ as the nonconformity score function, we define a quantile function over the calibration set as:
\begin{equation}\label{eq:quantile_value}
\textstyle	Q \left(1-\alpha; R_{\text{calib}} \right) = \text{inf} \left\{ r^* : \left( \frac{1}{\left|R_{\text{calib}} \right|} \sum_{r_i \in R_{\text{calib}}} \mathbb{I}_{r_i \leq r^*}\right) \geq {1-\alpha} \right\},
\end{equation}
where $\mathbb{I}$ is an indicator function (i.e., $\mathbb{I}=1$ if $r_i \leq r^*$; 0 otherwise). In essence, the term $\frac{1}{\left|R_{\text{calib}} \right|} \sum_{r_i \in R_{\text{calib}}} \mathbb{I}_{r_i \leq r^*}$ quantifies the proportion of samples satisfying $r_i \leq r^*$, so that $Q(1-\alpha;  R_{\text{calib}})$ identifies the smallest $r^*$ such that the proportion of samples in the calibration set with nonconformity score no more than $r^*$ is at least $1-\alpha$. In this manner, CP guarantees that the constructed prediction region is statistically valid under the condition of data exchangeability~\citep{shafer2008tutorial}. Mathematically, the prediction region for a given test data $\bm{X}_j  \in D_\text{test}$ is constructed as 
\begin{equation}
\textstyle   \Gamma^{1-\alpha}\left(\bm{X}_j\right):=\left[\hat{\bm{Y}}_j-Q\left(1-\alpha; R_{\text{calib}}\right), \; \hat{\bm{Y}}_j+Q\left(1-\alpha; R_{\text{calib}}\right)\right], \quad \forall \bm{X}_j \in D_{\text{test}}.
\end{equation}

However, as node-wise traffic data exhibits temporal dependency, CP fundamentally loses its validity due to the loss of data exchangeability in time-series data. To address this, we propose CPST to guarantee the validity of CP using temporal quantile adjustments \citep{lin2022conformal}. CPST introduces a variable $\hat{\delta}$ to dynamically adjust the quantile $\hat{\alpha} = \alpha - \hat{\delta}$ to query based on the calibration set. If the prediction region from CP in the calibration set does not meet the $1 - \alpha$ significance level, $\hat{\delta}$ is adjusted to align empirical coverage with the target. For node $i$ at time step $T_{c+1}$ in the test set, node-wise conformal prediction in a time series context involves the following two steps:
\begin{enumerate}
	\item Estimate $\hat{r}_{i, T_{c+1}}$ that reveals the quantile to query for node $i$ at time $T_{c+1}$: 
    \begin{equation} \label{eq:quantile}
        \begin{array}{c}
        \textstyle  \hat{r}_{i, T_{c+1}} = Q^{-1}\left(\overline{\epsilon}_{i, T_c}; \{\overline{\epsilon}_{j,T_c}\}_{j=1}^N \right), 
        \end{array}
    \end{equation}
    where $\overline{\epsilon}_{i,T_c} = \sum_{t' = 1}^{T_c}(\frac{\eta}{T_c}|y_{i, t'} - \hat{y}_{i, t'}| + \sum_{k \in \Omega(i)} \frac{\zeta}{T_c}|y_{k, t'} - \hat{y}_{k, t'}|) \beta^{T_c - t'}$ is the weighted residual (i.e., the nonconformity score) associated with node $i$ and its neighboring nodes $\Omega(i)$ up to time $T_c$, $\beta$ is the rate of exponential decay controlling the impact of preceding data on model prediction, $\eta$ and $\zeta$ ($\eta + \zeta = 1$) are parameters to adjust the relative importance of the residual associated with node $i$ and its neighboring nodes to enhance the robustness of conformal prediction; $Q^{-1}(\cdot)$ is an inverse quantile function that converts a given probability into the corresponding quantile. The goal is to find $\overline{\epsilon}_{i, T_c}$'s position in the set $\{\overline{\epsilon}_{j,T_c}\}_{j=1}^N$ to estimate the quantile $\hat{r}_{i, T_{c+1}}$.
	
	\item Given the estimated quantile $\hat{r}_{i,T_{c+1}}$ and preset coverage $1-\alpha$, compute $\hat{\delta}_{i,T_{c+1}}$ by an adjustment function $g$. That is, $\hat{\delta}_{i,T_{c+1}} = g(\hat{r}_{i,T_{c+1}}; \alpha)$ and
    \begin{equation}
      \textstyle  g\left(\hat{r}_{i,T_{c+1}} ; \alpha \right) 
        = \left\{\begin{array}{l}
		  C(\hat{r}_{i,T_{c+1}}-(1-\alpha)), \;\hat{r}_{i,T_{c+1}}<1-\alpha, \\
		\hat{r}_{i,T_{c+1}}-(1-\alpha), \; \hat{r}_{i,T_{c+1}} \geq 1-\alpha,
	\end{array}\right.
    \end{equation}
    where $C$ is a parameter to be tuned. When the estimated quantile $\hat{r}_{i,t+1}$ is less than $1-\alpha$, we increase the value of adjustment variable $\hat{\delta}_{i,t+1}$ to widen the original prediction region, and vice versa. 
    By doing so, the prediction region associated with $\hat{y}_{i,T_{c+1}}$ on the test set can be expressed as $[L_{i,T_{c+1}}, U_{i,T_{c+1}}] = [\hat{y}_{i, T_{c+1}}-\hat{v}_{i,T_{c+1}}, \hat{y}_{i, T_{c+1}}+\hat{v}_{i,T_{c+1}}]$, where $\hat{v}_{i,T_{c+1}} = Q(1-\hat{\alpha}_{i,T_{c+1}}; \{\overline{\epsilon}_{i,T_{c+1}}\}_{i=1}^N)$ is the adjusted quantile value of the nonconformity scores, with $\hat{\alpha}_{i,T_{c+1}} = \alpha - \hat{\delta}_{i,T_{c+1}}$. Note that when making prediction on the test set, the calibration set is dynamically updated. Specifically, when the predicted time becomes $T_{c+2}$, the actual and predicted values at time $T_{c+1}$ are added to the calibration set. Correspondingly, the first sample in the calibration set is removed to keep $n_{\text{calib}}$ unchanged; the updated calibration set is then used to generate nonconformity scores and update $\hat{\delta}_{i,T_{c+2}}$, and so on.
\end{enumerate}

\vspace{-0.5em}
\section{Computational Study}\label{sec:Computational Experiments}
We demonstrate the effectiveness of the proposed framework for traffic prediction using two real-world traffic datasets. The proposed method is first compared with several state-of-the-art models in prediction accuracy. Next, we assess the validity and efficiency of conformal prediction methods to showcase the superiority of the proposed conformal prediction method in traffic prediction.

\subsection{Dataset Description}
\vspace{-0.9em}
\begin{figure}[!ht]
	\centering
	\includegraphics[scale=0.55]{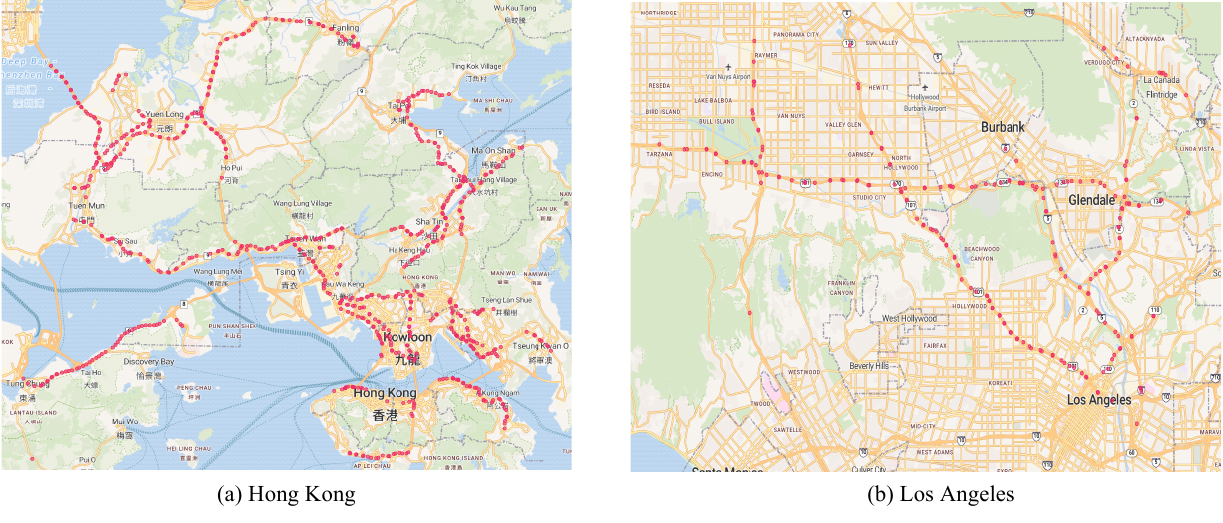}
        \vspace{-0.5em}
	\caption{Distribution of Traffic Detectors in the HK and METR-LA Datasets}
	\label{fig:7}
\end{figure}
\vspace{-1.7em}

The first real-world traffic dataset is the Hong Kong traffic dataset (referred to as the HK dataset hereafter). As a part of the ``smart mobility" initiatives, the Hong Kong Transport Department installed 700 traffic detectors (see Figure~\ref{fig:7}(a) for their distribution) on strategic routes and major roads across Hong Kong. The detectors record real-time traffic information, including traffic volume, traffic speed, and road occupancy. This study uses traffic data collected from July 1, 2023 to October 31, 2023, spanning a duration of 121 days. Due to data incompletion from 83 sensors, data from the remaining 617 detectors are used for performance study. Each detector samples data every 30 seconds, and the raw data is aggregated into 5-minute time interval, resulting in 35,335 samples per detector and a total of 21.8 million samples. The target variable in the HK dataset is traffic volume, indicating the number of vehicles on the roads in a given 5-minute window. The traffic network topology is derived from adjacency relationships extracted from OpenStreetMap.

The second dataset, the METR-LA dataset, is a widely used public dataset encompassing traffic speed data collected from 207 detectors (see Figure~\ref{fig:7}(b) for their distribution) in Los Angeles County, USA~\citep{jagadish2014big}. This dataset records traffic speed statistics over four months, from March 1, 2012 to June 30, 2012, with a 5-minute sampling frequency, resulting in 34,272 samples per detector. The road network's adjacency matrix is provided by~\citet{li2018diffusion}. In the METR-LA dataset, traffic speed is the target variable. In both datasets, traffic states are predicted every 5 minutes, with a prediction horizon ranging from 5 minutes (1 step) to 60 minutes (12 steps).

\vspace{-0.5em}
\subsection{Experiment Setup}\label{sec:experiment_setup}
\vspace{-1.8em}
\begin{table}[!ht]\scriptsize
    \caption{Configuration of the Tuned Model}
    \centering
    \begin{tabular}{cccccc}
    \toprule 
    Name & Value & Name &  Value & Name & Value\\ 
    \midrule
    Dimension of embedding layer & 32  & Kernel size of TCN &$(1,2)$ &  Graph convolution layer & 32 \\
    Embedding dimension of adaptive graph matrix & $(617,10)$ & Prediction layer & {$[512,256,12]$} & Learning rate & $e^{-3}$ \\
    Batch size & 64 & Epoch & 100  & Optimizer & Adam  \\
    Weight decay & $e^{-4}$ & Dropout rate & 0.3 & &\\
    \bottomrule
    \end{tabular}
    \label{tab:1}
\end{table}
\vspace{-2em}

The Hong Kong traffic dataset is split into training, validation, and test sets with a ratio of 6:2:2. Dates for each set are: training (July 1 to September 12), validation (September 12 to October 7), and test (October 7 to October 31). The METR-LA dataset is divided into training, validation, and test sets with a ratio of 7:1:2. Dates for each set are: training (March 1 to May 23), validation (May 23 to June 4), and test (June 4 to June 27). Note that the validation set serves a dual role: on the one hand, it is used to tune hyperparameters to prevent model overfitting; on the other hand, it serves as a calibration set for quantile query in building prediction regions for the test set. Deep learning models are trained using the PyTorch framework on a NVIDIA GeForce RTX 4090 GPU. Model validation indicates that the best-performing model consists of 4 layers, as summarized in Table~\ref{tab:1}. During inference, historical data from the previous 12 time steps are used to predict the next 12 time steps. A lag value of $P = 1$ is used to balance the spatial relationship between time-lagged and contemporaneous causal graphs. For conformal prediction, the parameters are $\eta = 0.95$, $\zeta = 0.05$, and $\beta = 0.9$ for the HK dataset, and $\eta = 0.99$, $\zeta = 0.01$, and $\beta = 0.9$ for the METR-LA dataset.

Three metrics\textemdash mean absolute errors (MAE), root mean squared errors (RMSE), and mean absolute percentage errors (MAPE) are used to compare the  deterministic prediction models:
\begin{equation}
    \begin{array}{cc}
    \textstyle \text{MAE} = \frac{1}{N \times H} \sum_{i=1}^{N} \sum_{t=1}^{H} \left| y_{i,t} - \hat{y}_{i,t} \right|,  \\
    \textstyle \text{RMSE} = \sqrt{\frac{1}{N\times H} \sum_{i=1}^{N} \sum_{t=1}^{H} \left( y_{i,t} - \hat{y}_{i,t} \right)^2}, \\
    \textstyle \text{MAPE} = \frac{1}{N \times H} \sum_{i=1}^{N} \sum_{t=1}^{H} \left| \frac{y_{i,t} - \hat{y}_{i,t}}{y_{i,t}} \right| \times 100\%,
    \end{array}
\end{equation}
where $N$ indicates the number of nodes in the road network, $H$ is the number of look-ahead time steps, $y_{i,t}$ and $\hat{y}_{i,t}$ refer to the ground truth and the predicted value, respectively. 

Moreover, we adopt two prevalent metrics\textemdash coverage and efficiency\textemdash defined below to assess the performance of conformal prediction for node-wise traffic prediction~\citep{shafer2008tutorial}:
\begin{equation}
	\begin{array}{c}
	\textstyle \text{Coverage}=\frac{1}{H} \sum_{t = 1}^{H} \mathbb{I}_{y_{i,t} \in[L_{i,t}, U_{i,t}]} , \forall i \in \{1,2, \cdots, N\}, \\ 
	\textstyle \text{Efficiency}=\frac{1}{H} \sum_{t = 1}^{H}\left(U_{i,t}-L_{i,t}\right), \forall i \in \{1,2, \cdots, N\},
	\end{array}
	\label{eqn:all-lines}
\end{equation}
where $U_{i,t}$ and $L_{i,t}$ are the upper and lower bounds produced by conformal prediction, respectively. 


\vspace{-0.5em}
\subsection{Causal Relationship Analysis}\label{sec:causal_relationship_analysis}

\vspace{-1.3em}
\begin{table}[!ht]\scriptsize
	\centering
	\caption{Summary on the Characteristics of Intra-slice and Inter-slice Causal Graphs}
	\begin{tabularx}{\textwidth}{c *{4}{>{\centering\arraybackslash}X}}
		\hline 
		Dataset & \multicolumn{2}{c}{HK} & \multicolumn{2}{c}{METR-LA} \\
		\hline
		Graph type & Intra-slice causal graph & Inter-slice causal graph & Intra-slice causal graph & Inter-slice causal graph \\
		\hline
		Number of nodes & 617 & 617 & 207 & 207 \\
		Number of edges & 2831 & 5078 & 364 & 470 \\
		Graph diameter & 5 & 5 & 7 & 9 \\
		Average shortest path & 2.78 & 2.56 & 0.128 & 0.807 \\
		Average degree & 10.48 & 16.81 & 3.51 & 4.54 \\
		\hline
	\end{tabularx}
	\label{tab:2}
\end{table}
\vspace{-2em}

To demonstrate the learned inter-slice and intra-slice causal graphs, we selected a one-week data for each dataset: July 1–7, 2023 for the HK dataset and March 1–7, 2012 for the METR-LA dataset. Table~\ref{tab:2} summarizes the basic characteristics of the discovered causal graphs. The graph diameter, calculated as the longest shortest path between any two nodes, provides insight into the maximum distance information throughout the graph. The average shortest path, computed by averaging the shortest paths between all node pairs, reflects the average distance between nodes and indicates overall connectivity. The average degree, the mean number of edges per node, measures node connectivity, with higher values indicating denser connections. Several key observations can be made from Table~\ref{tab:2}. First, both intra-slice and inter-slice causal graphs contain the same number of nodes for each dataset. The inter-slice causal graph, however, has more edges than the intra-slice graph because it accounts for the time-lagged causal effects of a node on itself. Furthermore, the inter-slice causal graph is more complex than the intra-slice graph. It has a higher average degree, indicating denser connectivity, while the intra-slice graph shows relatively sparse connectivity. Additionally, the causal graphs of the HK dataset have higher average shortest paths and degrees compared to the METR-LA dataset, suggesting that causal relationships in Hong Kong's traffic network have a stronger impact and longer propagation paths. Visualizations of the causal graphs for selected areas from both datasets are available in Section \ref{supp:causal-graph} of the Online Supplement.

\subsection{Performance Comparison}
In this study, we adopt two groups of methods as baselines to compare against our proposed method in traffic prediction and uncertainty estimation. In traffic prediction, we consider five deterministic models: DCRNN~\citep{li2018diffusion}, Graph WaveNet \citep{wu2019graph}, ASTGNN \citep{guo2021learning}, DDSTGCN \citep{sun2022dual}, and STCGAT \citep{zhao2022stcgat}. In uncertainty estimation, we consider four methods: SCP \citep{lei2018distribution}, CFRNN \citep{stankeviciute2021conformal}, TQA-B and TQA-E \citep{lin2022conformal}, see Table~\ref{tab:comparison} in the Online Supplement for detailed descriptions.

\vspace{-1.25em}
\begin{table}[!ht]\scriptsize
	\centering 
	\begin{threeparttable}
		\caption{Performance Comparison of Deterministic Model Prediction on the Test Set}
		\begin{tabular}{ccccccccc}
			\toprule
			\multirow{2}{*}{Dataset} & \multirow{2}{*}{Prediction Horizon} & \multirow{2}{*}{Metric} & \multicolumn{6}{c}{Model} \\
			\cmidrule{4-9}          &       &       & DCRNN & Graph WaveNet & ASTGNN & DDSTGCN & STCGAT & CASTMGCN \\
			\midrule    
			\multirow{12}[2]{*}{HK} & \multirow{3}[1]{*}{3 (15 mins)} & MAE   & 18.21 & 14.36 & 14.02 & 14.04 & 13.96 & \textbf{13.82} \\
			&       & RMSE  & 28.59 & 22.75 & 23.07 & 22.17 & 23    & \textbf{21.83} \\
			&       & MAPE  & 30.97\% & 25\%  & 25.81\% & 26.14\% & 25.30\% & \textbf{25.11\%} \\
			\cline{2-9}
			& \multirow{3}[0]{*}{6 (30 mins)} & MAE   & 19.75 & 15.36 & 14.95 & 14.67 & 15.09 & \textbf{14.42} \\
			&       & RMSE  & 31.08 & 24.55 & 25.17 & 23.37 & 25.3  & \textbf{22.95} \\
			&       & MAPE  & 33.49\% & 26.94\% & 27.04\% & \textbf{26\%}  & 26.81\% & 26.68\% \\
			\cline{2-9}
			& \multirow{3}[0]{*}{12 (60 mins)} & MAE   & 24    & 17.46 & 16.71 & 15.9  & 16.65 & \textbf{15.26} \\
			&       & RMSE  & 37.58 & 28.05 & 28.23 & 25.37 & 28.07 & \textbf{24.23} \\
			&       & MAPE  & 41.02\% & 31.26\% & 30.58\% & 29.40\% & \textbf{28.94}\% & 29.40\% \\
			\cline{2-9}
			& \multirow{3}[1]{*}{Average} & MAE   & 16.71 & 15.55 & 15.05 & 14.76 & 15.1  & \textbf{14.39} \\
			&       & RMSE  & 28.15 & 24.81 & 25.2  & 23.46 & 25.28 & \textbf{22.78} \\
			&       & MAPE  & 31.09\% & 27.40\% & 0.28  & 27.31\% & 27\%  & \textbf{26.53\%} \\
			\midrule
			\multirow{12}[2]{*}{METR-LA} & \multirow{3}[1]{*}{3 (15 mins)} & MAE   & 2.98  & 2.78  & 2.77  & 2.74  & 2.75  & \textbf{2.69} \\
			&       & RMSE  & 5.88  & 5.37  & 5.37  & 5.27  & 5.31  & \textbf{5.18} \\
			&       & MAPE  & 7.97\% & 7.42\% & 7.5\% & 7.16\% & 7.14\% & \textbf{7\%} \\
			\cline{2-9}
			& \multirow{3}[0]{*}{6 (30 mins)} & MAE   & 3.58  & 3.18  & 3.17  & 3.12  & 3.13  & \textbf{3.09} \\
			&       & RMSE  & 7.25  & 6.49\% & 6.48  & 6.3036 & 6.37  & \textbf{6.29} \\
			&       & MAPE  & 10.21\% & 9.07\% & 8.95\% & 8.57\% & 8.56\% & \textbf{8.43\%} \\
			\cline{2-9}
			& \multirow{3}[0]{*}{12 (60 mins)} & MAE   & 3.9   & 3.64  & 3.6   & 3.58 & 3.59 & \textbf{3.49} \\
			&       & RMSE  & 7.92  & 7.63  & 7.58  & 7.39  & 7.51  & \textbf{7.33} \\
			&       & MAPE  & 11.5\% & 10.84\% & 10.39\% & 10.14\% & 10.02\% & \textbf{9.95\%} \\
			\cline{2-9}
			& \multirow{3}[1]{*}{Average} & MAE   & 3.57  & 3.13  & 3.12  & 3.09  & 3.10  & \textbf{3.05} \\
			&       & RMSE  & 7.14  & 6.33  & 6.32  & 6.16  &  6.24 & \textbf{6.14} \\
			&       & MAPE  & 9.4\% & 8.86\% & 8.74\% & 8.4\% & 8.36\% & \textbf{8.27\%} \\
			\bottomrule
		\end{tabular}%
		\label{tab:performance_comparison}
	\end{threeparttable}
\end{table}%
\vspace{-1.5em}

Table~\ref{tab:performance_comparison} reports the prediction performance of all considered models on the test set. The proposed model consistently outperforms others across various metrics and prediction horizons on both datasets. Compared to DDSTGCN, our approach significantly reduces MAE, RMSE, and MAPE, highlighting the advantage of using multiple causal graphs to capture spatial dependencies. DCRNN, which combines diffusion convolution with recurrent neural networks, underperforms due to its inability to account for spatial dependencies from exogenous factors. Graph WaveNet and STCGAT, with their adaptive adjacency matrices, show better performance than DCRNN. ASTGNN, based on the self-attention mechanism, performs similarly to Graph WaveNet but requires substantial training time for large-scale road networks due to its multi-head attention structure. The computational costs of these methods are reported in Table~\ref{tab:6} in the Online Supplement.

\begin{figure}[!ht]
	\centering
	\includegraphics[scale=0.305]{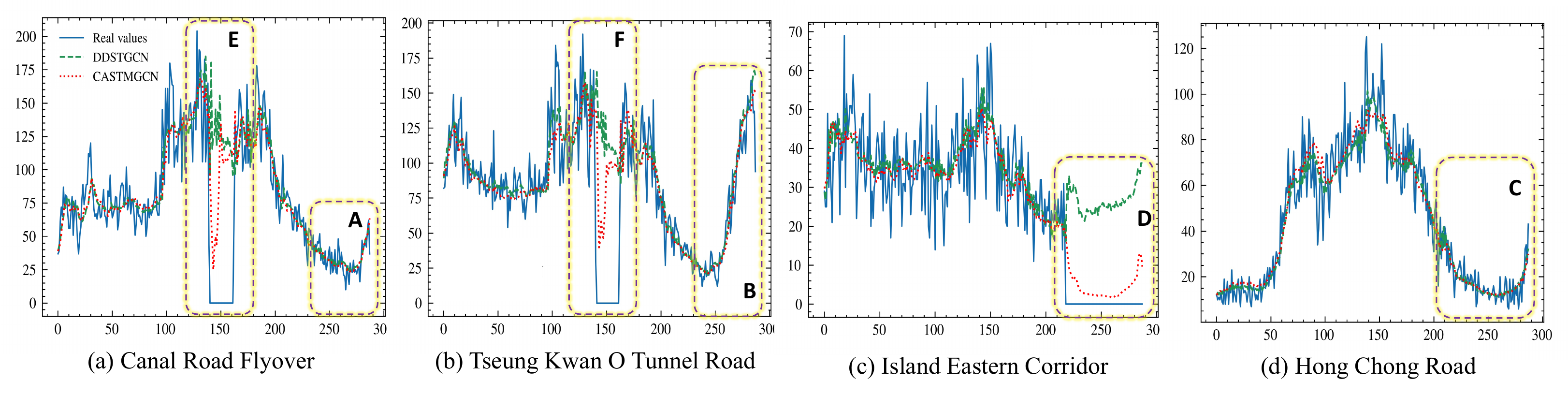}
        \vspace{-2.5em}
	\caption{Prediction Performance Comparison of CASTMGCN and DDSTGCN over Several Road Segments}
	\label{fig:8}
\end{figure}

Figure~\ref{fig:8} compares the prediction of CASTMGCN with the suboptimal model DDSTGCN over several road segments in Hong Kong. Panels (a)-(d) contrast model prediction and actual traffic flow over four locations: a flyover in Hong Kong Island, a main tunnel in Sai Kung District, an East District corridor near an underwater tunnel, and a main road in Kowloon. CASTMGCN surpasses DDSTGCN in the accuracy consistently. Both models capture general the overall traffic flow trends well, as seen in panels (a), (b), and (d), depicting traffic flow increases or decreases in regions A, B, and C. However, DDSTGCN struggles with abrupt changes in traffic flow, as observed in regions E and F in panels (a) and (b), and near-zero traffic volume in region D in panel (c). In contrast, CASTMGCN handles extreme scenarios and uncertainties, such as congestion and fluctuating demand, more robustly. Panel (d) highlights the complexity of major traffic roads, where CASTMGCN's performance, though slightly lower than in suburban areas (panel (b)), reflects the challenge of modeling traffic dynamics. 

\vspace{-1.5em}
\vspace{-2em}

\vspace{-1.3em}
\begin{table}[!ht]\scriptsize
	\centering
	\begin{threeparttable}
	\caption{Coverage and Efficiency for One-hour-ahead Traffic Prediction on the Test Set ($1-\alpha=0.9$)}
		\begin{tabular}{wl{16mm}p{12mm}wr{16mm}wr{16mm}wr{16mm}wr{16mm}wr{16mm}wr{16mm}}
			\toprule
			\multirow{2}{*}{Dataset} & \multicolumn{2}{c}{\multirow{2}{*}{Metric}} & \multicolumn{5}{c}{Method} \\
			\cmidrule{4-8}          & \multicolumn{2}{c}{} &  SCP    &   CFRNN & TQA-B & TQA-E & CPST \\
			\midrule
			\multirow{5}{*}{HK} & \multicolumn{2}{c}{Coverage} & 18.89\% & 89.15\% & 90.22\% & 92.18\% & 90.14\% \\
			\cline{2-8}
			& \multirow{4}{*}{Efficiency} & mean  & 20.81 & 69.54 & 71.14 &    Inf   & 70.92 \\
			&       & std   & 8.66  & 30.50  & 31.23 &    Inf   & 30.95 \\
			&       & max   & 50.93 & 178.18 & 183.78 &   Inf    & 176.80 \\
			&       & min   & 2.33  & 8.29  & 8.42  &      Inf & 8.07 \\
			\midrule
			\multirow{5}{*}{METR-LA} & \multicolumn{2}{c}{Coverage} &   44.41\%   & \textcolor[rgb]{.063,  .071,  .078}{84.22\%} & 88.18\% & 
			\textcolor[rgb]{.063,  .071,  .078}{93.46\%} &  93.22\% \\
			\cline{2-8}
			& \multirow{4}{*}{Efficiency} & mean  &    3.26   & 29.32 & 37.69 &     Inf  & 38.97 \\
			&       & std   &   3.92    & 26.05 & 37.33 &    Inf   & 39.16 \\
			&       & max   &    60.88   & 137.25 & 140.33 &   Inf    &  156.41\\
			&       & min   &     1.02  & 7.67  & 6.64  & 4.54  & 6.64 \\
			\bottomrule
		\end{tabular}%
		\label{tab:4}%
	\end{threeparttable}
\end{table}
\vspace{-2em}

Next, we compare our proposed method against several common conformal prediction methods using the two traffic datasets. Table~\ref{tab:4} summarizes the empirical coverage and efficiency of prediction regions on the test set targeting a coverage of 90\% ($\alpha = 0.1$). Our method achieves the narrowest prediction region among all the methods that are statistically valid. Though the prediction region of SCP has the smallest width, its empirical coverage with a value of 18.89\% and 44.41\% in the HK and METR-LA datasets, respectively, is distant away from the target coverage due to the non-exchangeability of time series data. Unlike SCP, TQA-E satisfies the target coverage, but it comes at the cost of producing a prediction region of an infinite width. 

\vspace{-1.2em}
\begin{figure}[!ht]
	\centering
	\includegraphics[width=0.82\linewidth]{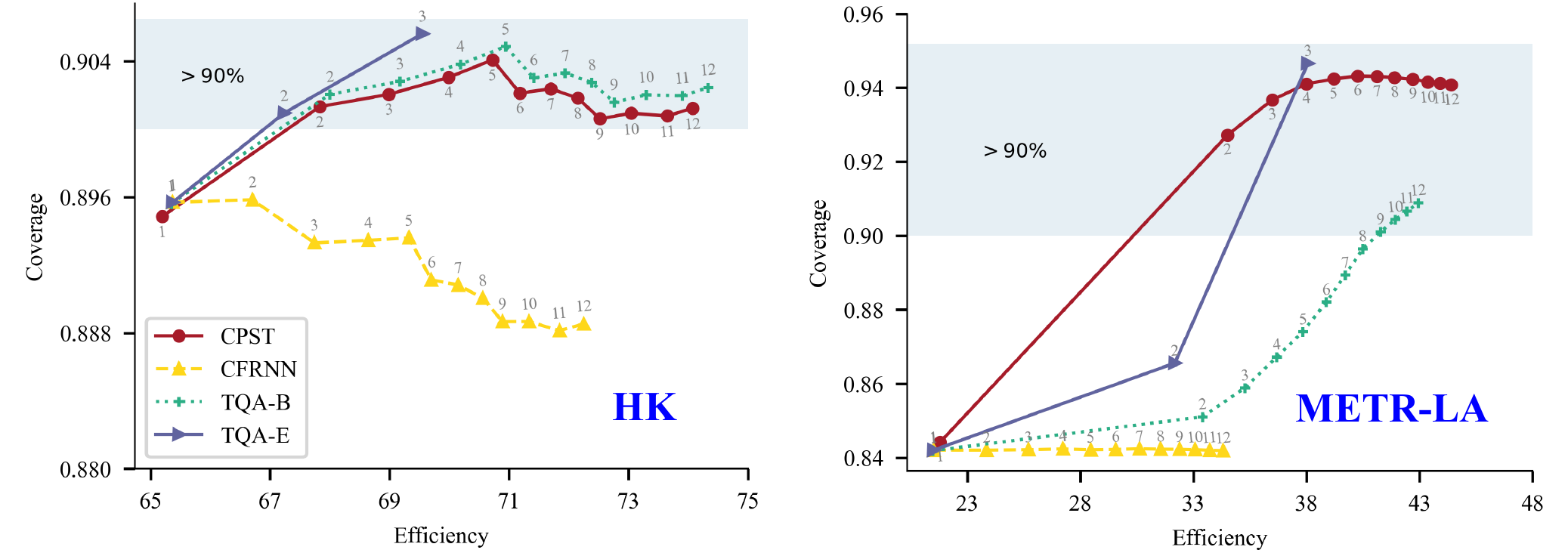}
        \vspace{-0.6em}
	\caption{Efficiency and Coverage of Several Conformal Prediction Methods on the Test Set over Varying Time Steps}
	\label{fig:15}
\end{figure}
\vspace{-1.8em}

Figure~\ref{fig:15} shows the mean coverage and efficiency of CFRNN, TQA-B, TQA-E, and CPST averaged over all the nodes for two traffic datasets with the number of prediction step increasing from 1 to 12. The text around each marker indicates the corresponding number of look-ahead time steps. The prediction region generated by all methods widens (reflected by increasing efficiency) as the number of prediction steps grows. For CPST and TQA-B, the coverage shows an upward trend as the number of prediction steps increases, exceeding 90\% after two steps. Notably, CPST does not reach the preset 90\% coverage in the first step due to the lack of historical records for quantile adjustment. After the first step, CPST's coverage stabilizes around the preset 90\%. CFRNN achieves the lowest prediction width at the cost of sacrificing coverage. TQA-E generates multiple prediction regions with infinite width after three time steps to satisfy the prescribed coverage, and no values are thus reported for TQA-E beyond this point. Overall, CPST achieves the highest efficiency among all the models while meeting the target coverage. In the Online Supplement (Section~\ref{supp:UQ}), we visualize the quantified uncertainty for randomly selected road segments in Hong Kong and prediction uncertainty across all time periods in both Hong Kong and Los Angeles. We also perform a comprehensive ablation study to illustrate the role of each considered graph, see Section~\ref{sec:ablation_study} in the Online Supplement for more details.

\vspace{-0.5em}
\section{Conclusion}\label{sec:Conclusion}
Accurate and reliable prediction is a paramount demand across a broad array of applications. In this paper, we take traffic prediction as an exemplary application to showcase a CASTMGCN framework developed for generating accurate and reliable traffic predictions over road network. In CASTMGCN, several heterogeneous graphs characterizing different aspects of network-wide traffic distributions, including causal graph, spatio-temporal dependencies, and other implicit traffic dependencies, are encoded into neural network to inform the modeling and representation learning of traffic patterns manifested in the observed traffic data. The proposed end-to-end learning framework not only provides a smart way to learn the often-overlooked traffic dependencies that are difficult to model explicitly, but also offers a convenient interface for fusing complex heterogeneous graphs in a data-driven manner. On this basis, we further develop a dedicated conformal prediction approach for spatio-temporal traffic data to quantify the uncertainty in node-wise traffic predictions. The statistically meaningful uncertainty estimation fosters development of robust measures to improve road network efficiency. Studies on two large traffic datasets suggest that the proposed method outperforms all baseline models in prediction accuracy. Besides, our CP method is not only statistically valid, but also generates more efficient prediction region than other CP approaches. 

In the future, the following research directions need to be further investigated. First, we will delve into the causal relationships among road segments to identify traffic bottlenecks by uncovering the causal propagation of congestion over the traffic network. Second, simulation platform (e.g., SUMO) can be leveraged to mimic real-world traffic situations. By implementing various control strategies in the simulation environment, we can identify the best strategy for improving the road network efficiency. Third, more data need to be collected around the nodes of high prediction uncertainty to further understand the complex traffic patterns at these locations. Finally, the proposed approach can be applied to other spatio-temporal modelling problems, such as air pollution, infectious disease spread, to help improve the accuracy and reliability of ML predictions in these contexts. 

\vspace{-0.7em}

\bibliographystyle{informs2014} 
\bibliography{ref} 





  



\ECSwitch

\ECHead{Online Supplement to ``Causally-Aware Spatio-Temporal Multi-Graph Convolution Network for Accurate and Reliable Traffic Prediction''}

\section{Comparisons with State-of-the-Art Methods}
Table~\ref{tab:overview} below summarizes key components of each relevant study across several dimensions reviewed earlier in Section~\ref{sec:literature_review}, underscoring current research gaps: neglect of implicit traffic patterns, lack of provable guarantees on the statistical validity of estimated uncertainty, and reliance on simplified assumptions about prior distributions in Bayesian uncertainty estimation.

\begin{table}[!ht]\scriptsize
	\centering
	\caption{Comparison of Our Proposed Methodology with the Existing Literature Relevant to Traffic Prediction}
	\begin{threeparttable}
	\begin{tabular}{p{2cm}|C{2.2cm}|C{2cm}|C{2.5cm}|C{2cm}|C{2.3cm}|C{1.6cm}}
			\hline
			\textbf{Study} & \textbf{Means of spatio-temporal modeling} & \textbf{Incorporation of causal structure?} & \textbf{Uncertainty quantification method}  & \textbf{Assumption on probability distribution} & \textbf{Statistical validity guarantee on estimated uncertainty?} & \textbf{Traffic variables of interest}\\
			\hline
			\citet{li2018diffusion}   & DCRNN  & $\times$ & $\times$ & N/A\tnote{1} & N/A & Traffic speed\\
			\hline
			\citet{jin2024spatial}   &  1D TCN and attention mechanism   &  $\times$ &  Bayesian approach with negative log-likelihood loss & Gaussian distribution& $\times$ & Traffic speed and flow\\
			\hline
			\citet{wu2023adaptive} & DNN &
			$\times$ & Quantile regression and conformal prediction  & Distribution free & \checkmark & Traffic speed and flow\\
			\hline
			\citet{qian2023towards, qian2023uncertainty}  & GCN and GRU  & $\times$ & Monte Carlo and conformal Calibration & Gaussian distribution &  \checkmark & Traffic flow\\
			\hline
			\citet{sengupta2024bayesian} &LSTM & $\times$ & Bayesian approach & Gaussian distribution & $\times$ & Traffic flow \\
			\hline
			\citet{lin2023dynamic}   &  Multi-head scaled dot product, GRU and GCN & \checkmark &  $\times$ & N/A & N/A & Traffic speed \\
			\hline
			\citet{luan2022traffic} & Dynamic GCN & \checkmark & $\times$ & N/A & N/A & Traffic speed \\
			\hline
			\textbf{Our method} & MGCN and Gated-ATCN  & \checkmark  & Conformal prediction &Distribution free & \checkmark & Traffic speed and flow\\
			\hline
		\end{tabular}
		\label{tab:overview}
		\begin{tablenotes}
			\item[1] N/A: Assumption on probability distribution and discussion on statistical validity are not applicable if the model does not estimate the prediction uncertainty. 
		\end{tablenotes}
	\end{threeparttable}
\end{table}

\section{Notation List}
For the sake of convenience, the mathematical notations used in this paper are summarized in Table~\ref{tab:notation}. 
 \begin{table}[!ht]\footnotesize 
	\centering
	\caption{List of Mathematical Notations } 
	\begin{threeparttable} 
		\begin{tblr}{Q[l, m, 3cm]Q[l, m, 13cm]}
			\hline
			Notation & Description \\ 
			\hline 
			$M/H/P$ & Historical/Predicted/Time-lagged time steps\\
			$\bm{X}_M$ & Historical traffic variables of the entire road network over $M$ time steps\\
			$\bm{x}_t$ &  Traffic state over the $N$ road segments at time step $t$\\
			$\bm{Y}_H$ & Traffic states for the entire road network over the next $H$ steps\\ 
			$V/E/\bm{A}$ & Set of nodes/Set of edges/Adjacency matrix\\
			$T_M$	&Time information over $M$ time steps\\
			$\bm{C}$/ $\bm{C}^{ij}$ & Contemporaneous effect/Contemporaneous effect of node $i$ on node $j$ at time step $t$\\
			$\bm{A}_t^{\circ}$  &Time-lagged causal matrices from time step $t-1$ to $t$\\
			$a_{t}^{ij}$ & Time-lagged effect of node $i$ on node $j$ at time step $t$ \\
			$\text{Att}_M$ & Masked attention matrix \\
			$\bm{Q},\bm{K},\bm{V}$ & Query, key, and value components \\
			$\hat{\bm{A}}$ & Auxiliary adaptive adjacency matrix\\
			$\lambda_{\bm{C}}$, $\lambda_{A^{\circ}}$, $\frac{\rho}{2}$  & Regularization coefficients\\
			$\alpha_L$ & Lagrange multiplier\\
            $\rho$ & Penalty parameter\\			$\bm{X}_M^e/\bm{X}_M^a/\bm{X}_c/\bm{X}_s$ & 
			Output of the embedding/masked self-attention/gated-TCN/MGCN  \\
			$\bm{W}_* $&   Weighted matrices \\
			$b_*$ & Biases\\
			$1- \alpha$ & Target coverage\\
			$\overline{\epsilon}_{i,t}$ & Nonconformity score of node $i$ at time step $t$ \\
			$\Omega(i)$ & Neighboring nodes of node $i$\\
			$\hat{\delta}_{i,t}$ & Quantile adjustment variable of node $i$ at time step $t$ \\ 
			$\hat{r}_{i,t}$ & Estimated quantile of node $i$ at time step $t$\\
			$\hat{v}_{i,t}$ & Quantile value of node $i$ at time step $t$\\
			$R_{\text{calib}}$ & Nonconformality scores for samples in the calibration set \\
   \hline
		\end{tblr}
	\end{threeparttable}
	\label{tab:notation}
\end{table}



\section{Visualization of Discovered Causal Relationships} \label{supp:causal-graph}

\begin{figure}[!ht]
	\centering
	\includegraphics[width=0.7\linewidth]{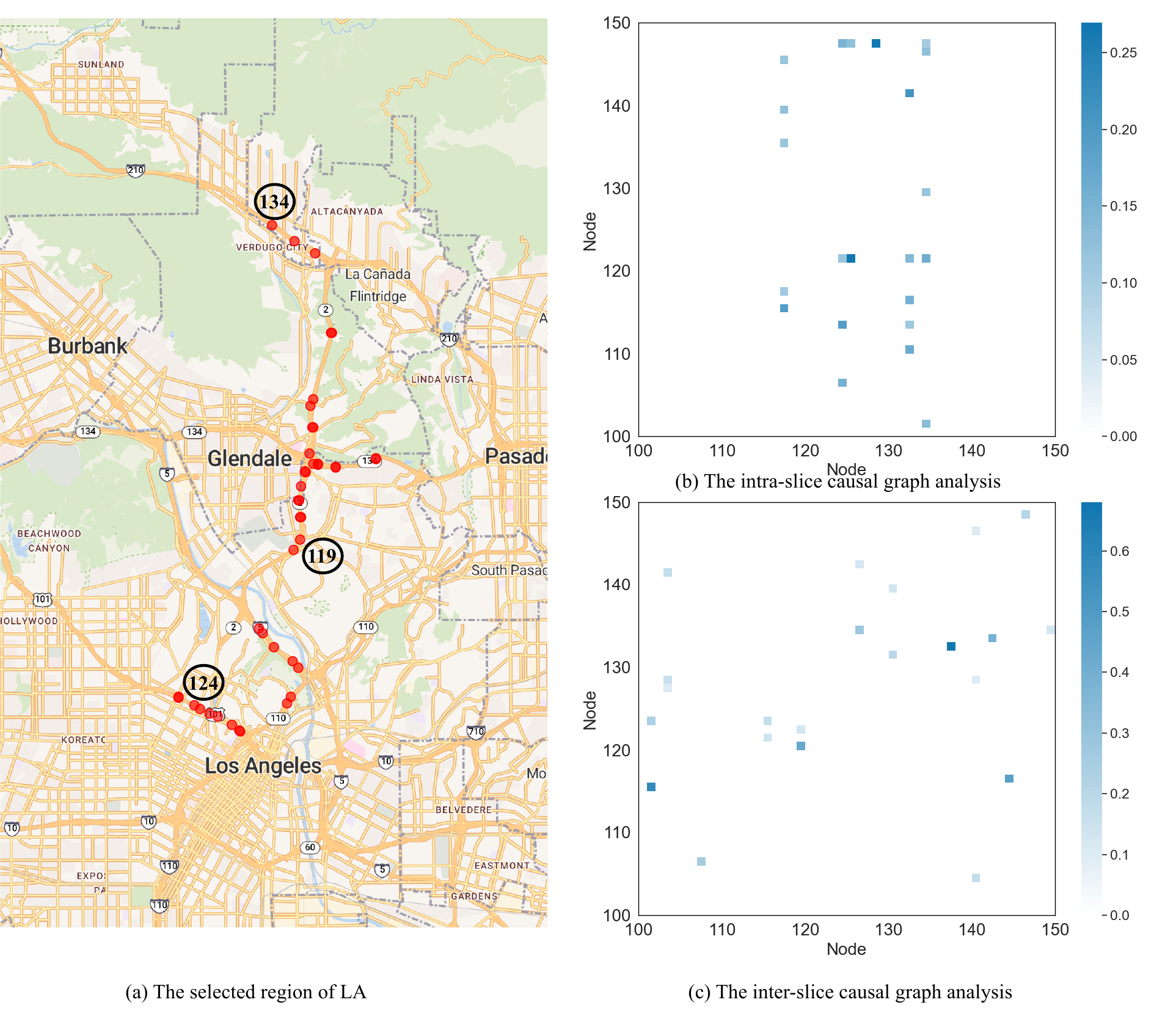}
	\caption{Causal Graph Analysis of a Selected Region in the LA Dataset. Note: The selected region contains 50 nodes.}
	\label{fig:causal_la}
\end{figure}

\begin{figure}[!ht]
	\centering
	\includegraphics[width=0.7\linewidth]{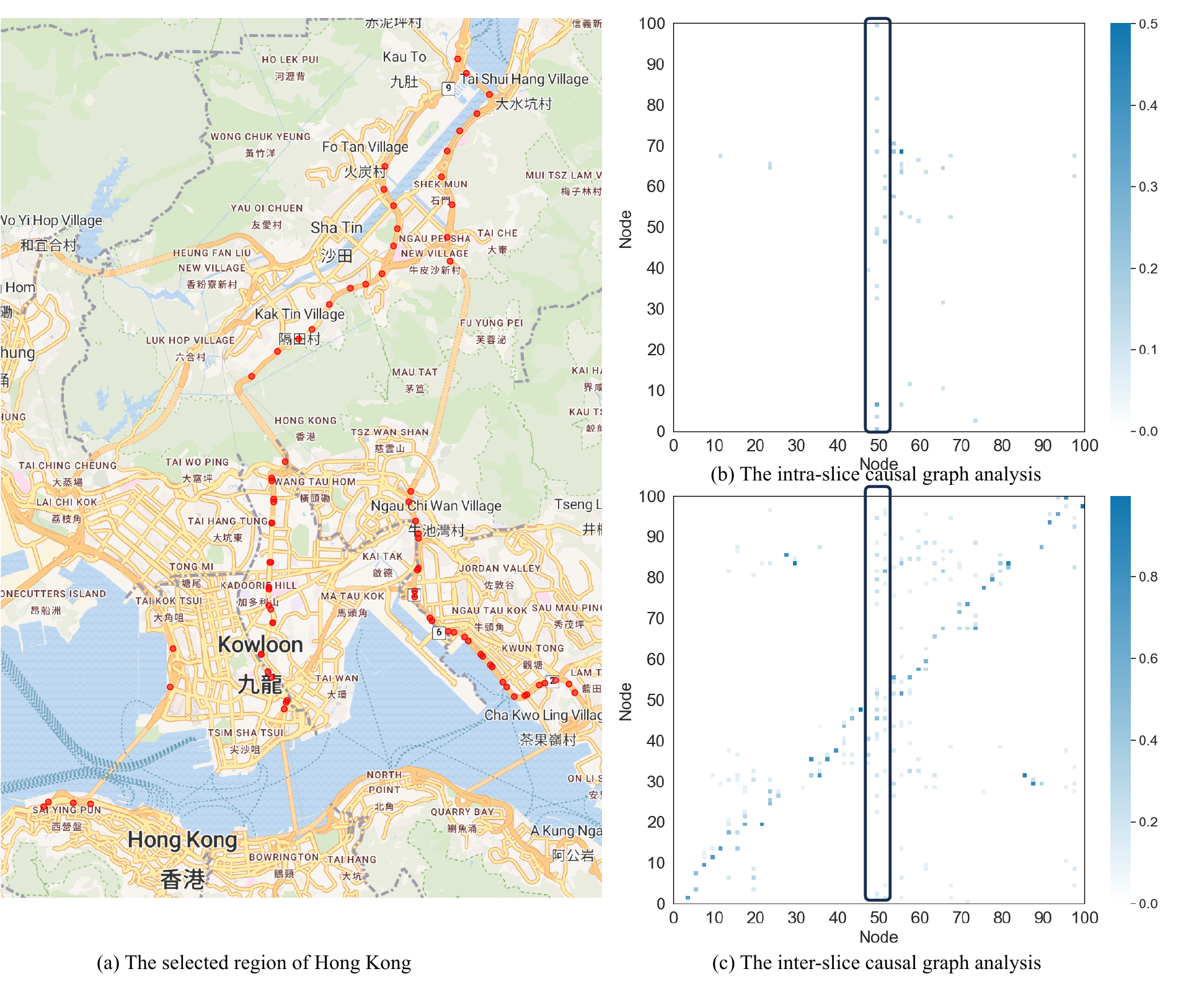}
	\caption{Causal Graph Analysis of a Selected Region in the HK dataset. Note: The selected region contains 100 nodes.}
	\label{fig:12}
\end{figure}

To illustrate the discovered causal relationships, we select representative areas from the two datasets for causal graph visualization. Figures~\ref{fig:causal_la}(a) and \ref{fig:12}(a) display the selected areas in LA and HK, respectively, and Figures~\ref{fig:causal_la}(b), \ref{fig:causal_la}(c), \ref{fig:12}(b), and \ref{fig:12}(c) show heatmaps for inter-slice and intra-slice causalities in the two considered areas. In LA, the intra-slice dependencies of several nodes, including 119, 124, and 134, are significantly higher than others, indicating that they are strongly influenced by other nodes. On the map, we can see that the three nodes are located at the end of road segments, meaning that these nodes have neighbors only on one side. Regarding the HK dataset, the values of diagonal elements in the inter-slice dependency are substantially higher than others, indicating that the time-lagged causal effect of a node on itself is much stronger than that from other nodes. In addition, Figure~\ref{fig:12}(b) shows that some columns, particularly around columns 49-50, have significantly higher values than others, suggesting localized areas of strong effect. Moreover, the values in columns 49-50 in Figure~\ref{fig:12}(c) are slightly higher, indicating that this area is more affected by other road segments.

\begin{figure}[!ht]
	\centering
	\includegraphics[width=0.7\linewidth]{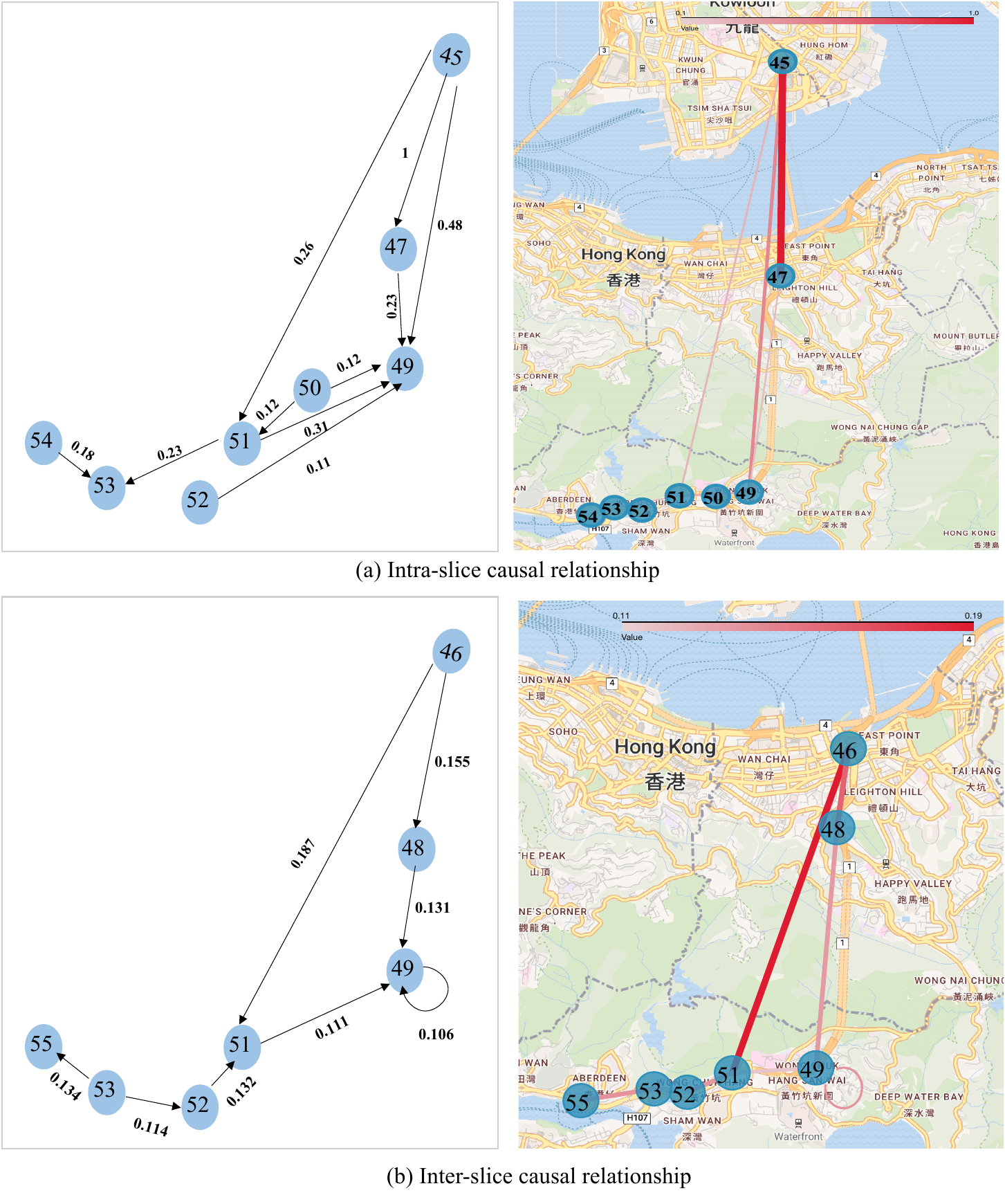}
	\caption{The Inter-slice and Intra-slice Causal Relationships of the 49th Node and Its Neighboring Nodes}
	\label{fig:13}
\end{figure}

Figure~\ref{fig:13} further visualizes the causal graph for nodes 45 to 55 on the geographical map in Hong Kong. Note that edges with causal effect less than 0.1 are excluded to guarantee visualization quality. Node 49 is located near the toll plaza of the Aberdeen Tunnel. The toll plaza is a complex multi-intersection and thus susceptible to the traffic conditions of other neighboring road segments. Panel (b) depicts the time-lagged causal relationships of nodes beyond adjacent nodes. Specifically, node 51 is affected not only by its neighbor 52, but also by the distant node 46. Node 46 is located near one side of the Cross Harbor Tunnel, and the associated traffic detector records a large number of vehicles passing through the tunnel to reach node 51. Therefore, there is a strong causal relationship between the two nodes. Panel (a) depicts the contemporaneous causal relationships. We can observe that there are more edges in Panel (b) than in Panel (a). Node 49 with a degree of 5 is more significantly affected by other road segments nearby. In fact, node 49 is located at a critical traffic hub and carries a larger traffic volume, thus deserving increased attention in traffic management. In addition, we can easily identify causal effect propagation phenomena from Panel (a), such as $45 \rightarrow 47 \rightarrow 49 $ and $45 \rightarrow 51 \rightarrow 53$. These discovered causal chains of events enable identifying traffic bottlenecks leading to congestion in traffic network and their root causes.

\section{Additional Details on the Graph Processing Operator $g(a)$}\label{sec:g_a}
In MGCN, we use the normalized Laplacian operator to process adjacency matrix $\bm{A}$ for smoothing the feature representation and retaining the unique characteristics of each node. Mathematically, for a given node $i$, when $k = 1$, \eqref{eq:MGCN} becomes $g(\bm{A})\bm{X}_c^{(i)}=\bm{X}_c^{(i)}-\sum_{j\in \Omega(i)} \frac{1}{\sqrt{d_i}}\bm{A}_{ij}\frac{1}{\sqrt{d_j}}\bm{X}_c^{(j)}$~\citep{yang2019masked}, where $\Omega(i)$ is the set of neighbors around node $i$ in graph $\bm{A}$, $d_i$ is the degree of node $i$, and $\bm{A}_{ij}$ is the weight of the edge connecting nodes $i$ and $j$. This operation considers both the node's own features and its neighbors' features. It smooths feature representation by eliminating sharp local variations, and also captures the difference between the node's own features and its neighboring nodes' features. Besides, we use the normalized asymmetric operator to process the learned causal graphs $(\bm{A}^{\circ}, \bm{C})$ for balancing each neighbor's contribution. Mathematically, this operator is defined as $g(a) = \bm{D}_{\text{out}}^{-1} a$, $a \in \{\bm{A}^{\circ},\bm{C}\}$~\citep{yang2019masked}. For $k=1$, the convolution operation on feature $\bm{X}_c^i$ is given by  $g(a)\bm{X}^{(i)}\bm{W}_k = \sum_{j \in \Omega(i)} \frac{1}{d_i}a_{ij}\bm{X}_c^{(j)} \bm{W}_k$, where $a_{ij}$ indicates the causal strength from node $i$ to node $j$, $d_i$ is the out-degree of node $i$, and $\bm{D}_{\text{out}}$ is out-degree of graph $a$. This operation computes the weighted sum of node's neighbors' feature, balancing their contributions to the overall feature representation. For the graph processing operator $g(a)$, $g(a)^k$ (i.e., the matrix product of $k$ copies of $g(a)$) integrates spatial effects into the Gated-ATCN feature representation. When $a = \bm{A}$, the $(i, j)$-th entry of $a^k$  indicates the number of ways to travel between nodes $i$ and $j$ in $k$  moves. If $a = \bm{A}^{\circ}$, it represents the cumulative effect of time-lagged causality between nodes $i$ and $j$ via exactly $k$ moves. The product of $g(a)^k$ and $\bm{X}_c$ makes the node-wise feature representation $\bm{X}_c$ spatially-aware. Naturally, a smaller (resp. larger) value of $k$ induces MGCN to capture more local (resp. global) spatial information.

\section{Additional Details on Performance Comparison}
Table~\ref{tab:comparison} presents detailed description on each baseline model. Specifically, we consider five deterministic models\textemdash DCRNN, Graph WaveNet, ASTGNN, DDSTGCN and STCGAT\textemdash for traffic prediction, and four methods\textemdash SCP, CFRNN, TQA-B and TQA-E\textemdash for uncertainty estimation.
	
\begin{table}[!ht]\scriptsize 
	\centering 
	\caption{Baseline Model Descriptions} 
	\begin{threeparttable} 
		\begin{tblr}{Q[l, m, 2cm]|Q[l, m, 5cm]|Q[l, m,2.6cm]|Q[l, m, 5.2cm]}
			\hline
			\textbf{Deterministic models} & \textbf{Description} & \textbf{CP models} & \textbf{Description} \\ 
			\hline
			DCRNN \citep{li2018diffusion} & Models traffic flow as a diffusion process using a directed graph and constructs a diffusion convolutional recurrent neural network to predict traffic flow. & SCP~\citep{lei2018distribution} & A direct application of split conformal prediction. \\ \hline
			Graph WaveNet \citep{wu2019graph} & Develops an adaptive adjacency matrix and utilizes node embeddings to learn spatial dependency during model training. It combines graph convolution and dilated causal convolution, simultaneously, to make spatio-temporal predictions. & CFRNN~\citep{stankeviciute2021conformal} & Bonferroni correction is used to ensure effective coverage of the entire time horizon (all $T$ steps) in time series forecasting. With Bonferroni correction, CFRNN avoids generating infinitely wide prediction intervals when performing correct split conformal prediction. \\ \hline
			ASTGNN \citep{guo2021learning} & Extends the transformer model by incorporating a trend-aware attention mechanism and accounting for the periodicity and spatial heterogeneity of traffic data. & TQA-B~\citep{lin2022conformal} & A conformal prediction method based on time series data that improves coverage by adjusting the quantile level of prediction intervals and exhibits a certain efficiency in the width of prediction regions. \\ \hline
			DDSTGCN \citep{sun2022dual} & Proposes a dual dynamic modeling approach to capturing both the dynamic properties of nodes and the dynamic spatio-temporal features of edges. & TQA-E~\citep{lin2022conformal} & A conformal prediction method based on time series data that improves coverage by adjusting the quantile level of prediction intervals, which achieves improved coverage by using prediction intervals of infinite width. \\ \hline
			STCGAT \citep{zhao2022stcgat} & Utilizes adaptive spatial adjacency subgraphs and efficient causal temporal correlation components to make accurate traffic prediction. &  &  \\ \hline
		\end{tblr} 
		\label{tab:comparison}
	\end{threeparttable} 
\end{table}

Table~\ref{tab:6} compares the computational costs of CASTMGCN against other deterministic prediction models. For both traffic datasets, the training and inference time needed by our model is among the smallest. Among all the models, Graph WaveNet is the most computationally efficient, followed by DDSTGCN and CASTMGCN. By contrast, ASTGNN, DCRNN, and STCGAT consume the most amount of computation time. Considering the superior performance attained by CASTMGCN, a slight increase in the computation time is generally acceptable. In practical applications, the difference in the computation costs (particularly the inference times) between CASTMGCN and Graph WaveNet is insignificant.

\begin{table}[!ht]
	\centering\small
	\caption{Computational Time of CASTMGCN and Other Deterministic Prediction Models}
	\begin{tabular}{lrccrr}
		\cline{1-6}
		\multirow{2}{*}{Model} & \multicolumn{2}{c}{HK dataset} &                      &        \multicolumn{2}{c}{MTER-LA dataset}              \\ \cline{2-6}
		& Training (sec/epoch) & Inference (sec) &                      &     Training (sec/epoch) & Inference (sec)                 \\ \hline
		DCRNN                  & 449.31               & 72.27           &                      &   157.00	&35.20                   \\
		Graph WaveNet          & 85.81                & 10.11           &                      &       23.47&	0.74               \\
		ASTGNN                 & 220.70               & 32.35           &                      &   45.50	&22.98                 \\
		DDSTGCN                & 108.30               & 18.96           &                      &     51.26	&2.31                 \\
		STCGAT                 & 354.56               & 54.63           &                      &      76.55	&8.85                \\
		CASTMGCN (Proposed)                 & 137.14               & 16.31           & &24.99	&1.19 \\ \cline{1-6}
	\end{tabular}
	\label{tab:6}
\end{table}

\section{Visualization of Uncertainty Estimation} \label{supp:UQ}
\begin{figure}[!ht]
	\centering
	\includegraphics[width=0.85\linewidth]{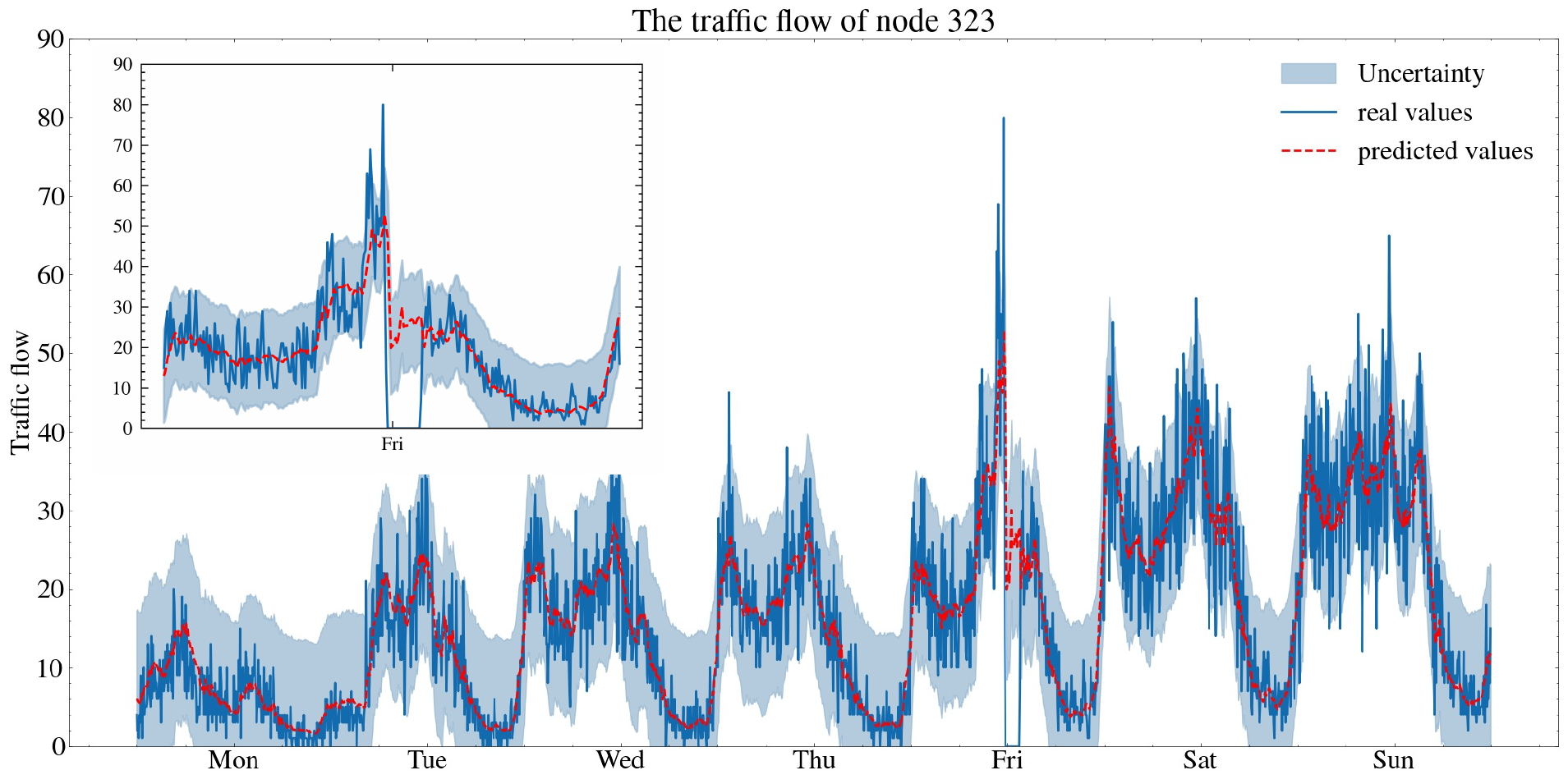}
	\caption{Uncertainty Quantification on Randomly Selected Nodes}
	\label{fig:9}
\end{figure}

Figure~\ref{fig:9} presents the ground truth, model prediction, and associated uncertainty for randomly selected road segments in the HK dataset over a one-week period in the test set. As can be observed, the predicted traffic flow generated by deterministic forecasting model generally align well with the patterns of traffic flow variation. However, as previously stated, challenges arise in extreme scenarios, such as traffic flow on Friday. This magnified area corresponds to data recorded by inductive loop detectors, and it relies on electromagnetic induction to record vehicles. During peak periods at time $t$, when vehicular volume surpasses road capacity, traffic congestion happens. This leads to a sharp decrease in the recorded traffic information occurs at time $t+1$, resulting in the manifestation of the phenomenon observed on Friday. Therefore, for such road segments, the estimation of uncertainty in the prediction results enhances the reliability of outcomes generated by deterministic models.

\subsection{Uncertainty Across All Time Periods}
\begin{figure}[!ht]
        \centering
	\subfigure[Hong Kong]{\includegraphics[width=0.455\textwidth]{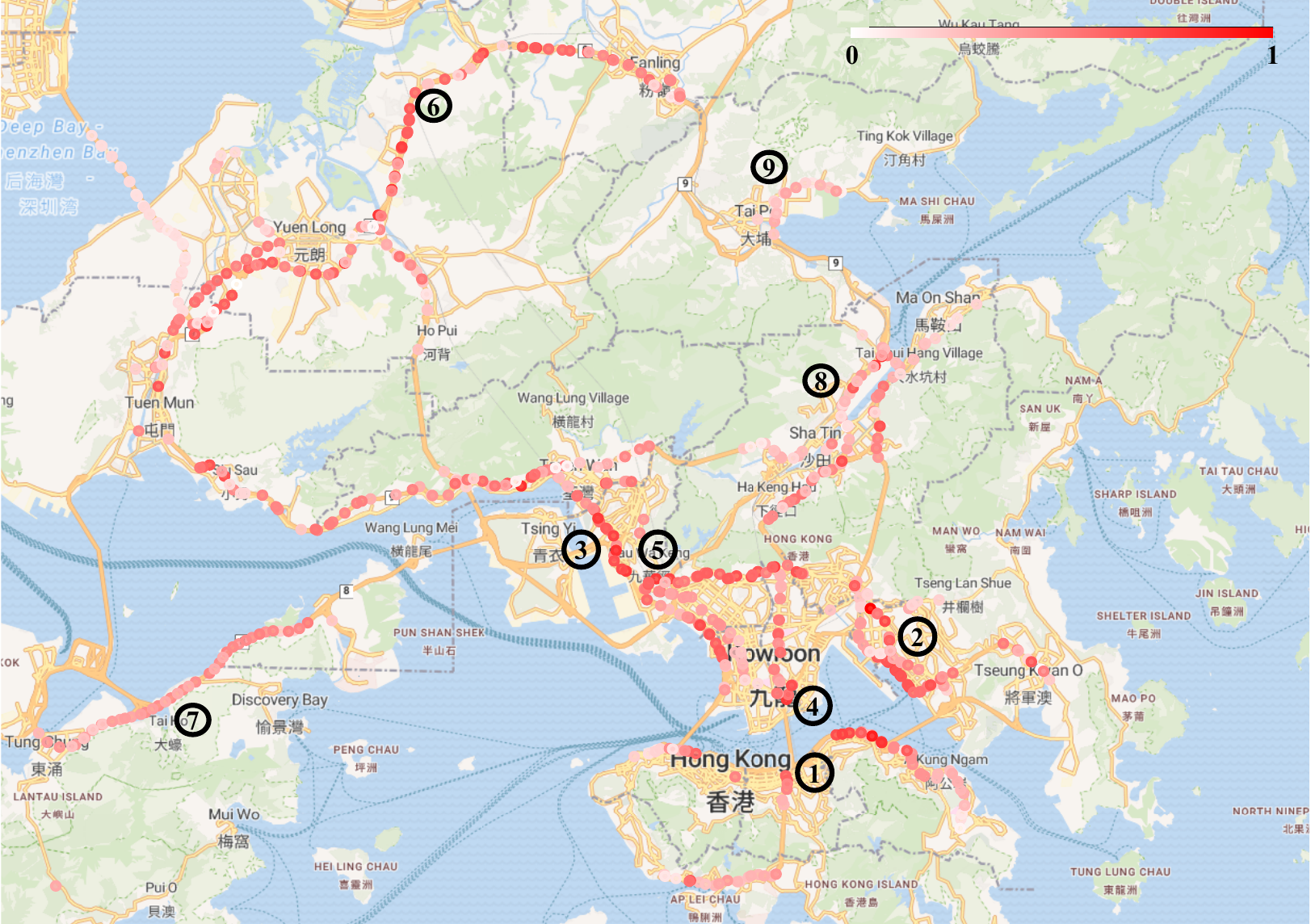}\label{fig:10}}
	\hspace{2mm}
	\subfigure[Los Angeles]{\includegraphics[width=0.436\textwidth]{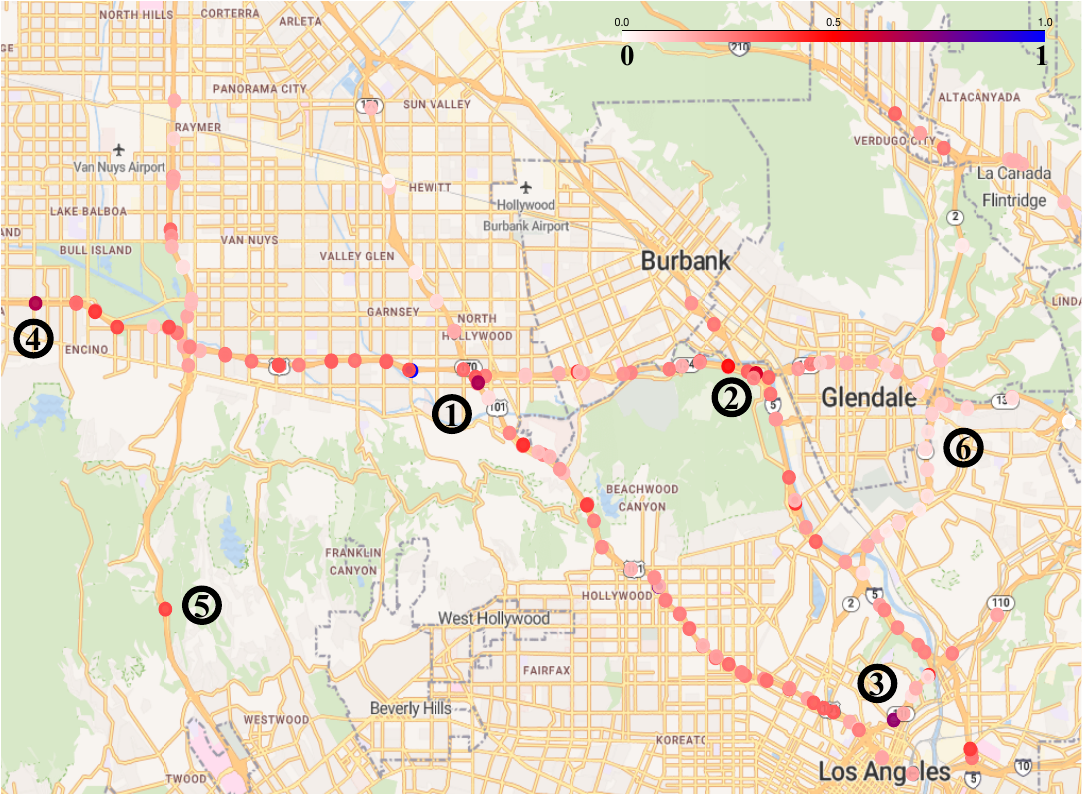}\label{fig:10_1}}
	\caption{Distribution of Uncertainty Averaged over all Time Periods}
	\label{fig:correlation_comparison}
\end{figure}

Figure~\ref{fig:correlation_comparison} visualizes the uncertainty of different road segments (with prediction intervals at a coverage of 90\%) on the test data for the HK and METR-LA datasets, respectively. Based on the segment-level uncertainty visualization, we draw the following findings:
\begin{enumerate}
	\item [(1)] Road segments with similar level of uncertainty exhibit clustering characteristics in large-scale traffic networks. For example, road segments in close proximity are often associated with similar level of uncertainty, as observed in \circled{\small3} and \circled{\small7} in Figure~\ref{fig:10}. However, this phenomenon is not pronounced in the METR-LA dataset.
	
	\item [(2)] Due to lower traffic volume and fewer external factors affecting traffic conditions, road segments located in the remote areas, such as the Tai Po Road connecting Tai Po to Sha Tin (annotated as \circled{\small8} and \circled{\small9} in Figure~\ref{fig:10}), the North Lantau Highway in Tsuen Wan District (marked as \circled{\small7} in Figure~\ref{fig:10}) and Ventura Freeway (marked as \circled{\small6} in Figure~\ref{fig:10_1}), exhibit lower uncertainty than other road segments in the vicinity of downtown area (marked as \circled{\small1}, \circled{\small2}, \circled{\small3}, \circled{\small4}, \circled{\small5} in Figure~\ref{fig:10} and \circled{\small1}, \circled{\small2} in Figure~\ref{fig:10_1}).
	
	\item [(3)] Notably, road segments close to the cross-harbour tunnel (marked as \circled{\small4} in Figure~\ref{fig:10}) and Eastern harbor crossing (marked as \circled{\small1} in Figure~\ref{fig:10_1}) show a higher prediction uncertainty. This is primarily attributed to the lack of sensors within the tunnels, thus resulting in a large fluctuation of traffic flow at both ends of the tunnel. In METR-LA, \circled{\small5} in Figure~\ref{fig:10_1} also exhibits a high uncertainty due to traffic speed variations caused by the lack of sensors in the neighboring area.
	
	\item [(4)]Road segments marked as \circled{\small3} and \circled{\small5} in Figure~\ref{fig:10} and \circled{\small1} and \circled{\small2} in Figure~\ref{fig:10_1} are located in the proximity of multiple intersections and thus exhibit higher uncertainty compared to other road segments. This is mainly attributed to the complexity and unpredictability of traffic states at the intersections. These intersections are potential bottlenecks in the traffic network, where traffic conditions are affected by a broad range of factors, such as intersection signal control, driving behavior, and traffic demand. Therefore, traffic conditions are highly uncertain in these areas.
	
	\item [(5)] It is worth noting that, despite being located on the outskirts, the road segment connecting Sha Po Village in Yuen Long District with San Tin Highway (marked as \circled{\small6} in Figure~\ref{fig:10}) exhibits a high uncertainty. This is because this road serves as a major backbone to connect Shenzhen Huanggang port via Xinshen Road and the Qing Shan Highway–San Tin section, thus facilitating travel to and from Lamma Island and other areas in San Tin. The uncertain demand at border crossings contributes to the uncertainty in predicting traffic states along this road segment.
\end{enumerate}

\subsection{Uncertainty Variation Over Different Time Periods}
\begin{figure}[!ht]
	\centering
	\includegraphics[width=0.9\linewidth]{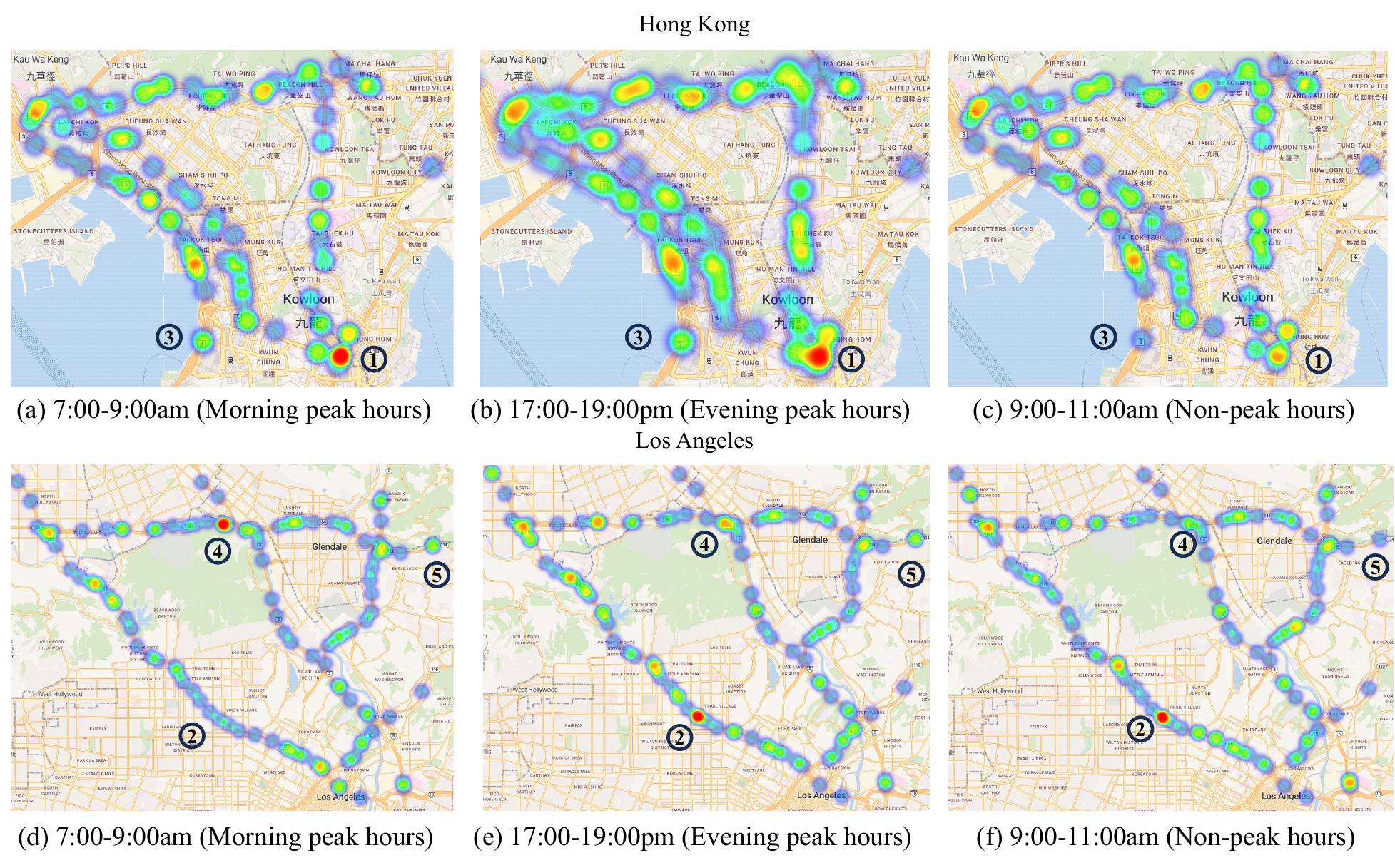}
	\caption{Variation of Uncertainty Associated with Traffic Prediction over Different Time Periods}
	\label{fig:16}
\end{figure}

Figure~\ref{fig:16} illustrates the variation of uncertainty across different time periods in both Hong Kong and Los Angeles. 
We can see that most road segments exhibit a high uncertainty during peak hours than off-peak periods. Certain specific road segments, such as those marked as \circled{\small1} and \circled{\small3} in Hong Kong and \circled{\small4} and \circled{\small5} in Los Angeles, exhibit significantly higher uncertainty during morning and evening peak hours compared to the off-peak hours. \circled{\small1} is a main road in Hung Hom, Kowloon, and it serves as one of the entry points to the Cross-Harbour Tunnel. \circled{\small3} is the Hoi Po Road that connects to the Western Harbour Crossing. Due to the complexity of road restrictions and tunnel conditions during morning and evening peak hours, the uncertainty associated with predicting the traffic flow is relatively higher. \circled{\small4} and \circled{\small5} are located at intersections and the traffic patterns are highly complex during morning and evening peak hours at intersections. As a result, deterministic models face challenges in accurately predicting the traffic flow at these locations. It is noteworthy that \circled{\small4} exhibits a relatively low uncertainty during the morning peak period, but higher uncertainty in the predictions during the evening peak and off-peak periods. This could be due to events or activities affecting traffic flow at that location, such as road construction, traffic accidents. Generally, quantifying uncertainty in deterministic traffic predictions is crucial as the traffic demand is constantly changing over time.
	
\section{Ablation Study}\label{sec:ablation_study}
\begin{table}[!ht]\scriptsize
    \centering
    \caption{Experimental Configuration of Ablation Study}
    \begin{tabular}{ccccc}
        \toprule 
            Model & Adjacency matrix &  
            Inter-slice causal graph &  
            Intra-slice causal graph &  
            Adaptive matrix \\
        \midrule 
        Model 1 & \checkmark & & & \\
        Model 2 & \checkmark & & & \checkmark \\
        Model 3 &  \checkmark & & \checkmark & \checkmark \\
        Model 4 & \checkmark & \checkmark & & \checkmark \\
        Model 5 & \checkmark & \checkmark & \checkmark & \\
        Model 6 & \checkmark & \checkmark & \checkmark & \checkmark \\
        \bottomrule
        \label{tab:5}
    \end{tabular}
\end{table}

\begin{figure}[!ht]
	\centering
	\includegraphics[width=0.95\linewidth]{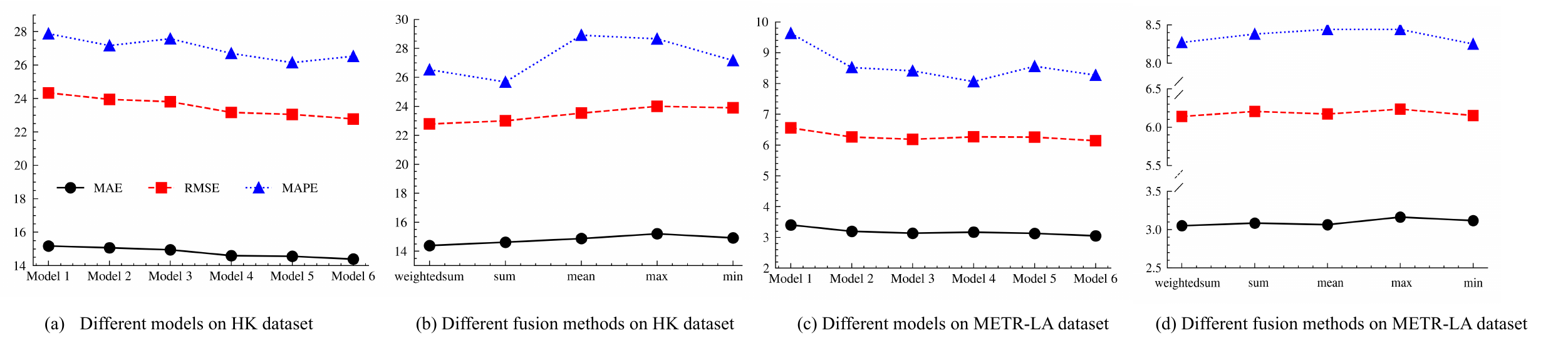}
	\caption{Results of Ablation Study. The value of MAPE is in percentage, and the percentage symbol is omitted.}
	\label{fig:11}
\end{figure}

In ablation study of the CASTMGCN model, we systematically evaluate the impact of its constituent modules on prediction performance. We consider six variants of the CASTMGCN model by incrementally adding each module to examine their effects (see Table~\ref{tab:5}). Figure~\ref{fig:11}(a) and (c) show MAE, RMSE, and MAPE metrics for a fixed prediction horizon of 12 steps on both datasets. Not surprisingly, the sixth model integrating all modules achieves the best performance, while the model relying solely on the adjacency matrix performs the worst. This highlights the significant contributions of both inter-slice and intra-slice causal graphs in enhancing prediction accuracy, underscoring their importance in guiding model learning.

In Section~\ref{sec:multi_graph_fusion}, we fuse several graphs capturing different aspects of spatial dependencies in the traffic network. Figure~\ref{fig:11}(b) and (d) compare the performance of five different fusion operations (weighted sum, sum, average, maximum, minimum) on the two datasets. The weighted sum operation achieves the best performance, suggesting that assigning varying weights to spatial features derived from multiple GCNs is more effective for accurately characterizing spatial dependencies. This underscores the importance of the adjacency matrix $\bm{A}$, inter-slice causal matrix $\bm{A}^{\circ}$, intra-slice causal matrix $\bm{C}$, and adaptive matrix $\hat{\bm{A}}$, each playing a crucial role in shaping spatially-aware feature representations. Conversely, the maximum and minimum operations perform poorly, as they do not account for the varying importance of each spatial feature in data fusion.

\end{document}